\documentclass{article}

\usepackage{arxiv}

\usepackage[utf8]{inputenc} 
\usepackage[T1]{fontenc}    
\usepackage{lmodern}        
\usepackage{hyperref}       
\usepackage{url}            
\usepackage{booktabs}       
\usepackage{amsfonts}       
\usepackage{nicefrac}       
\usepackage{microtype}      
\usepackage{lipsum}
\usepackage{graphicx}

\title{Surrogate Model Based Hyperparameter Tuning for Deep Learning
with SPOT}

\author{
    Thomas Bartz-Beielstein, Frederik Rehbach, Amrita Sen, and Martin
    Zaefferer
    \thanks{\url{https://www.spotseven.de}}
   \\
    Institute for Data Science, Engineering, and Analytics \\
    Technische Hochschule Köln \\
  5164 Gummersbach, Germany \\
  \texttt{\href{mailto:thomas.bartz-beielstein@th-koeln.de}{\nolinkurl{thomas.bartz-beielstein@th-koeln.de}}} \\
  }

\usepackage{color}
\usepackage{fancyvrb}

\DefineVerbatimEnvironment{Highlighting}{Verbatim}{commandchars=\\\{\}}
\usepackage{framed}
\definecolor{shadecolor}{RGB}{248,248,248}
\newenvironment{Shaded}{\begin{snugshade}}{\end{snugshade}}

\newcommand{\AttributeTok}[1]{\textcolor[rgb]{0.77,0.63,0.00}{#1}}

\newcommand{\ConstantTok}[1]{\textcolor[rgb]{0.00,0.00,0.00}{#1}}
\newcommand{\ControlFlowTok}[1]{\textcolor[rgb]{0.13,0.29,0.53}{\textbf{#1}}}

\newcommand{\DecValTok}[1]{\textcolor[rgb]{0.00,0.00,0.81}{#1}}

\newcommand{\FloatTok}[1]{\textcolor[rgb]{0.00,0.00,0.81}{#1}}
\newcommand{\FunctionTok}[1]{\textcolor[rgb]{0.00,0.00,0.00}{#1}}

\newcommand{\NormalTok}[1]{#1}

\newcommand{\OtherTok}[1]{\textcolor[rgb]{0.56,0.35,0.01}{#1}}

\newcommand{\SpecialCharTok}[1]{\textcolor[rgb]{0.00,0.00,0.00}{#1}}

\newcommand{\StringTok}[1]{\textcolor[rgb]{0.31,0.60,0.02}{#1}}

\usepackage{amsmath,amsfonts,amssymb,amsthm,url}
\usepackage{csquotes}
\usepackage{enumerate}
\usepackage{natbib}
\usepackage[nonumberlist]{glossaries}
\setcitestyle{square,aysep={},yysep={;}}
\usepackage{tabulary}
\usepackage{xspace}
\usepackage{centernot}

\glsdisablehyper
\makenoidxglossaries

\newtheorem{example}{Example}

\theoremstyle{definition}
\newtheorem{defn}{Definition}

\newcommand{\sgd}{\text{\textsc{SGD}}}
\newcommand{\momentum}{\text{\textsc{Momentum}}}
\newcommand{\nesterov}{\text{\textsc{Nesterov}}}
\newcommand{\adam}{\text{\textsc{Adam}}}
\newcommand{\nadam}{\text{\textsc{NAdam}}}
\newcommand{\rmsprop}{\text{\textsc{RMSProp}}}

\newcommand{\Xtest}{\ensuremath{X^{(\text{test})}}}
\newcommand{\Xtrain}{\ensuremath{X^{(\text{train})}}}
\newcommand{\Xtrainval}{\ensuremath{X^{(\text{train } \cup \text{ val})}}}
\newcommand{\Xval}{\ensuremath{X^{(\text{val})}}}
\newcommand{\psitest}{\ensuremath{\psi^{(\text{test})}}}
\newcommand{\psitrain}{\ensuremath{\psi^{(\text{train})}}}
\newcommand{\psival}{\ensuremath{\psi^{(\text{val})}}}
\newcommand{\psivalcv}{\ensuremath{\psi_{\text{CV}}^{(\text{val})}}}
\newcommand{\facctest}{\ensuremath{f_{\text{acc}}^{(\text{test})}}}
\newcommand{\facctrain}{\ensuremath{f_{\text{acc}}^{(\text{train})}}}
\newcommand{\faccval}{\ensuremath{f_{\text{acc}}^{(\text{val})}}}
\newcommand{\hatlambda}{\ensuremath{\hat{\lambda}}}

\newacronym{ADAM}{ADAM}{ADAptive Moment estimation algorithm}
\newacronym{AI}{AI}{Artificial Intelligence}
\newacronym{AID}{AID}{Automatic Interaction Detection}
\newacronym{API}{API}{Application Programming Interface}
\newacronym{AutoDL}{AutoDL}{Automated Deep Learning}
\newacronym{AutoHAS}{AutoHAS}{Automated Hyperparameter and Architecture Search}
\newacronym{AutoML}{AutoML}{Automated Machine Learning}
\newacronym{AutonoML}{AutonoML}{Autonomous Machine Learning}
\newacronym{BCE}{BCE}{Binary Cross Entropy}
\newacronym{BLAS}{BLAS}{Basic Linear Algebra Subprograms}
\newacronym{BLEU}{BLEU}{Bi-Lingual Evaluation Understudy}
\newacronym{BO}{BO}{Bayesian Optimization}
\newacronym{BOHB}{BOHB}{Bayesian Optimization HyperBand}
\newacronym{CAAI}{CAAI}{Cognitive Architecture for Artificial Intelligence}
\newacronym{CART}{CART}{Classification and Regression Trees}
\newacronym{CASH}{CASH}{Combined Algorithm Selection and Hyperparameter optimization}
\newacronym{CCE}{CCE}{Categorical Cross Entropy}
\newacronym{CIFARTEN}{CIFAR-10}{Canadian Institute for Advanced Research, 10 classes}
\newacronym{CNN}{CNN}{Convolutional Neural Network}
\newacronym{CV}{CV}{Cross Validation}
\newacronym{CRAN}{CRAN}{Comprehensive R Archive Network}
\newacronym{CRISP-DL}{CRISP-DL}{Cross-Industry Standard Process for Deep Learning}
\newacronym{CRISP-DM}{CRISP-DM}{Cross-Industry Standard Process for Data Mining}
\newacronym{DACE}{DACE}{Design and Analysis of Computer Experiments}
\newacronym{DeepOBS}{DeepOBS}{Deep Learning Optimizer Benchmark Suite}
\newacronym{DNN}{DNN}{Deep Neural Network}
\newacronym{DL}{DL}{Deep Learning}
\newacronym{DOE}{DOE}{Design of Experiments}
\newacronym[\glslongpluralkey=Evolution Strategies]{ES}{ES}{Evolution Strategy}
\newacronym{GB}{GB}{Gradient Boosting}
\newacronym{GBM}{GBM}{Gradient Boosting Model}
\newacronym{HB}{HB}{Hyperband}
\newacronym{HPO}{HPO}{Hyperparameter Optimization}
\newacronym{HPT}{HPT}{Hyperparameter Tuning}
\newacronym{IMDB}{IMDB}{Internet Movie Data Base}
\newacronym{MAE}{MAE}{Mean Absolute Error}
\newacronym{ML}{ML}{Machine Learning}
\newacronym{MNIST}{MNIST}{Modified National Institute of Standards and Technology}
\newacronym{MSE}{MSE}{Mean Squared Error}
\newacronym{NADAM}{NADAM}{Nesterov-accelerated Adaptive Moment Estimation}
\newacronym{NAS}{NAS}{Neural Architecture Search}
\newacronym{NLP}{NLP}{Natural Language Processing}
\newacronym{NN}{NN}{Neural Network}
\newacronym{OCBA}{OCBA}{Optimal Computing Budget Allocation}
\newacronym{R}{R}{R software environment for statistical computing and graphics}
\newacronym{RGB}{RGB}{Red, Green, and Blue color space}
\newacronym{RMSProp}{RMSProp}{Root Mean Square Propagation}
\newacronym{RNN}{RNN}{Recurrent Neural Network}
\newacronym{SGD}{SGD}{Stochastic Gradient Descent}
\newacronym{SMAC}{SMAC}{Sequential Model-Based Optimization for General Algorithm Configuration}
\newacronym{SMBO}{SMBO}{Surrogate Model Based Optimization}
\newacronym{SPOT}{SPOT}{Sequential Parameter Optimization Toolbox}
\newacronym{SPOTMisc}{SPOTMisc}{Sequential Parameter Optimization Toolbox -- Miscelleanous Functions}
\newacronym{SSQ}{SSQ}{Sum of Squares}
\newacronym{TF}{TF}{TensorFlow}
\newacronym{tfruns}{tfruns}{Training Run Tools for TensorFlow}
\newacronym{XGBoost}{XGBoost}{Extreme Gradient Boosting}

\DeclareMathOperator*{\argmin}{arg\,min}

\newcommand{\RbuildKriging}{\ensuremath{\operatorname{buildKriging}}\xspace}

\newcommand{\RlossMeanSquaredError}{\ensuremath{\operatorname{loss\_mean\_squared\_error()}}\xspace}
\newcommand{\RlossMeanAbsoluteError}{\ensuremath{\operatorname{loss\_mean\_absolute\_error()}}\xspace}
\newcommand{\RlossCategoricalCrossentropy}{\ensuremath{\operatorname{loss\_categorical\_crossentropy()}}\xspace}

\newcommand{\RoptimizerAdam}{\ensuremath{\operatorname{optimizer\_adam() }} }

\newcommand{\RoptimizerRmsprop}{\ensuremath{\operatorname{optimizer\_rmsprop() }} }

\begin{document}
\maketitle

\def\tightlist{}

\begin{abstract}
A surrogate model based hyperparameter tuning approach for deep learning
is presented. This article demonstrates how the architecture-level
parameters (hyperparameters) of deep learning models that were
implemented in Keras/tensorflow can be optimized. The implementation of
the tuning procedure is 100\% accessible from R, the software
environment for statistical computing. With a few lines of code,
existing R packages (tfruns and SPOT) can be combined to perform
hyperparameter tuning. An elementary hyperparameter tuning task (neural
network and the MNIST data) is used to exemplify this approach.
\end{abstract}


\section{Introduction}\label{sec:introduction}

\gls{DL} models require the specification of a set of architecture-level
parameters, which are called \emph{hyperparameters}. Hyperparameters are
to be distinguished from the \emph{parameters} of a model that are
optimized in the initial loop, e.g., during the training phase via
backpropagation. Hyperparameter values are determined before the model
is executed---they remain constant during model development and
execution whereas parameters are modified. We will consider \gls{HPT},
which is much more complicated and challenging than parameter
optimization (training the weights of a \gls{NN} model).

Typical questions regarding hyperparameters in \gls{DL} models are as
follows:

\begin{enumerate}
\item How many layers should be stacked? 
\item Which dropout rate should be used? 
\item How many filters (units) should be used in each layer? 
\item Which activation function should be used? 
\end{enumerate}

Empirical studies and benchmarking suites are available, but to date,
there is no comprehensive theory that adequately explains how to answer
these questions. Recently, \citet{Robe21a} presented a first attempt to
develop a \gls{DL}theory.

In real-world projects, \gls{DL} experts have gained profound knowledge
over time as to what reasonable hyperparameters are, i.e., \gls{HPT}
skills are developed. These skills are based on human expert and domain
knowledge and not on valid formal rules. Figure 1 in \citet{Kedz20a}
nicely illustrates how data scientists select models, specify metrics,
preprocess data, etc. \cite{chol18a} describe the situation as follows:

\begin{quote}
\enquote{If you want to get to the very limit of what can be achieved on a given task, 
you can’t be content with arbitrary [hyperparameter] choices made by a fallible human. 
Your initial decisions are almost always suboptimal, even if you have good intuition. 
You can refine your choices by tweaking them by hand and retraining the model repeatedly---that’s what 
machine-learning engineers and researchers spend most of their time doing.\\
But it shouldn’t be your job as a human to fiddle with hyperparameters all day---that is better left to a machine.}
\end{quote}

\gls{HPT} develops tools to explore the space of possible hyperparameter
configurations \emph{systematically}, in a structured way. For a given
space of hyperparameters \(\Lambda\), a \gls{DNN} model \(A\) with
hyperparameters \(\lambda\), training, validation, and testing data
\Xtrain, \Xval and \Xtest, respectively, a loss function \(L\), and a
hyperparameter response surface function \(\psi\), e.g., mean loss, the
basic \gls{HPT} process looks like
this\footnote{Symbols used in this study are summarized in Table \ref{tab:symbols}.}:

\begin{enumerate}[(HPT-1)]
\item Set $t=1$. Parameter selection (at iteration $t$). Choose a set of hyperparameters from the space of hyperparameters, 
$\lambda(t) \in \Lambda$.
\item \gls{DNN} model building. Build the corresponding \gls{DNN} model $A_{\lambda(t)}$.
\item \gls{DNN} model training and evaluation. Fit the model $A_{\lambda(t)}$ to the training data \Xtrain and measure the final performance, e.g., expected loss, on the validation data \Xval, i.e.,
\begin{equation}\label{eq:psival}
\psival = \frac{1}{| \Xval|} \sum_{x \in \Xval} L \left( x; A_{\lambda^{(t)}} (\Xtrain) \right),
\end{equation}
where $L$ denotes a loss function.
Under $k$-fold \gls{CV} the performance measure from Equation \ref{eq:psival} can be written as
\begin{equation}\label{eq:psivalcv}
\psivalcv = \frac{1}{k} \sum_{i=1}^k \frac{1}{| \Xval|} \sum_{x \in \Xval_i} L \left( x; A_{\lambda^{(t)}} (\Xtrain_i) \right),
\end{equation}
because the training and validation set partitions are build $k$ times.
\item Parameter update. The next set of hyperparameters to try, $\lambda(t+1)$, is chosen accordingly to minimize the performance, e.g., $\psival$.
\item Looping. Repeat until budget is exhausted.
\item Final evaluation of the best hyperparameter set $\lambda^{(*)}$ on test (or development) data \Xtest, i.e.,
measuring performance on the test (hold out) data  
\begin{equation}
\psitest = \frac{1}{| \Xtest|} \sum_{x \in \Xtest} L \left( x; A_{\lambda^{(*)}} (\Xtrainval) \right).
\end{equation}
\end{enumerate}

Essential for this process is the \gls{HPT} algorithm in (HPT-4) that
uses the validation performance to determine the next set of
hyperparameters to evaluate. Updating hyperparameters is extremely
challenging, because it requires creating and training a new model on a
dataset. And, the hyperparameter space \(\Lambda\) is not continuous or
differentiable, because it also includes discrete decisions. Standard
gradient methods are not applicable in \(\Lambda\). Instead,
gradient-free optimization techniques, e.g., pattern search or
\glspl{ES}, which sometimes are far less efficient than gradient
methods, are applied.

The following \gls{HPT} approaches are popular:

\begin{itemize}
\item manual search,
\item simple random search, i.e., choosing hyperparameters to evaluate at random, repeatedly,
\item grid and pattern search \citep{Meignan:2015vp} \citep{Torczon00} \citep{Tats16a},
\item model free algorithms, i.e., algorithms that do not explicitly make use of a model, e.g., \glspl{ES} \citep{Hans06a} \citep{Bart13j},
\item hyperband, i.e., a multi-armed bandit strategy that dynamically allocates resources 
to a set of random configurations and uses successive halving to stop poorly performing configurations\citep{Li16a}, 
\item \gls{SMBO} such as \gls{SPOT}, \citep{BLP05}, and \citep{bart21b}.\footnote{The acronym \gls{SMBO} originated in the 
engineering domain \citep{Book98a}, \citep{Mack07a}. It is also popular in the \gls{ML} community, where it stands for \emph{sequential model-based optimization}.
We will use the terms \emph{sequential model-based optimization} and \emph{surrrogate model-based optimization} synonymously.}
\end{itemize}

Manual search and grid search are probably the most popular algorithms
for \gls{HPT}. Interestingly, \cite{Berg12a} demonstrate empirically and
show theoretically that randomly chosen trials are more efficient for
\gls{HPT} than trials on a grid. Because their results are of practical
relevance, they are briefly summarized here: In grid search the set of
trials is formed by using every possible combination of values, grid
search suffers from the curse of dimensionality because the number of
joint values grows exponentially with the number of hyperparameters.

\begin{quote}
\enquote{For most data sets only a few of the hyperparameters really matter, 
but that different hyperparameters are important on different data sets. 
This phenomenon makes grid search a poor choice for configuring algorithms for new data sets}\citep{Berg12a}.
\end{quote}

Let \(\Psi\) denote the space of hyperparameter response functions (as
defined in the Appendix, see Definition \ref{def:surface}).
\cite{Berg12a} claim that random search is more efficient than grid
search because a hyperparameter response function \(\psi \in \Psi\)
usually has a low effective dimensionality; essentially, \(\psi\) is
more sensitive to changes in some dimensions than others
\citep{Calf97a}.

Due to its simplicity, it turns out in many situations, especially in
high-dimensional spaces, that random search is the best solution.
Hyperband should also mentioned in this context, although it can result
in a worse final performance than model-based approaches, because it
only samples configurations randomly and does not learn from previously
sampled configurations \citep{Li16a}. \cite{Berg12a} note that random
search can probably be improved by automating what manual search does,
i.e., using \gls{SMBO} approaches such as \gls{SPOT}.

\gls{HPT} is a powerful technique that is an absolute requirement to get
to state-of-the-art models on any real-world learning task, e.g.,
classification and regression. However, there are important issues to
keep in mind when doing \gls{HPT}: for example, validation-set
overfitting can occur, because hyperparameters are optimized based on
information derived from the validation data.

\cite{Falk18a} claim, that practical \gls{HPO} solutions should fulfill
the following requirements:

\begin{itemize}
\item strong anytime and final performance, 
\item effective use of parallel resources, 
\item scalability, as well as robustness and flexibility.
\end{itemize}

In the context of benchmarking, a treatment for these issues was
proposed by \cite{bart20gArxiv}. Although their recommendations (denoted
as (R-1) to (R-8)) were developed for benchmark studies in optimization,
they are also relevant for \gls{HPT}, because \gls{HPT} can be seen as a
special benchmarking variant.

\begin{enumerate}[(R-1)]
    \item Goals: what are the reasons for performing \gls{HPT}? 
    \item Problems: how to select suitable problems? Can surrogates accelerate the tuning?
    \item Algorithms: how to select a portfolio of \gls{DL} algorithms to be included in the \gls{HPT} study?  
    \item Performance: how to measure performance?  
    \item Analysis: how to evaluate results? 
    \item Design: how to set up a study, e.g., how many runs shall be performed?  
    \item Presentation: how to describe results? 
    \item Reproducibility: how to guarantee scientifically sound results and how to guarantee a lasting impact, e.g., in terms of comparability?  
\end{enumerate}

In addition to these recommendations, there are some specific issues
that are caused by the \gls{DL} setup. These will be discussed in Sec.
\ref{sec:discussion}.

Note, some authors used the terms \gls{HPT} and \gls{HPO} synonymously.
In the context of our analysis, these terms have different meanings:

\begin{description}
\item[\gls{HPO}] develops and applies methods to determine the best hyperparameters 
in an effective and efficient manner.
\item[\gls{HPT}] develops and applies methods that try to analyze the effects and interactions of 
hyperparameters to enable \emph{learning and understanding}.
\end{description}

This article proposes a \gls{HPT} approach based on \gls{SPOT} that
focuses on the following topics:

\begin{description}
\item[Limited Resources.] We focus on situations, where
limited computational resources are available.
This may be simply due the availability and cost of hardware, 
or because confidential data has to be processed strictly locally.
\item[Understanding.] In contrast to standard \gls{HPO} approaches,
\gls{SPOT} provides statistical tools for \emph{understanding}\/ hyperparameter importance
and interactions between several hyperparameters.
\item[Transparency and Explainability.] Understanding is a key tool for enabling 
transparency, e.g., quantifying the contribution of \gls{DL}  components (layers, activation functions, etc.).
\item[Reproducibility.] The software code used in this study is available in the open source \gls{R} package \gls{SPOT} via the \gls{CRAN}.
\gls{SPOT} is a well-established open-source software, that is maintained for more 
than 15 years \citep{BLP05}.
\end{description}

For sure, we are not seeking the overall best hyperparameter
configuration that results in a \gls{NN} which outperforms any other
\gls{NN} in every problem domain \citep{Wolp97a}. Results are specific
for one problem instance---their generalizibility to other problem
instances or even other problem domains is not self-evident and has to
be proven \citep{Haft16b}.

This paper is structured as follows: Section \ref{sec:materials}
describes materials and methods that were used for the experimental
setup. Experiments are described in Sec. \ref{sec:experiments}. Section
\ref{sec:results} presents results from a simple experiment. A
discussion is presented in Sec. \ref{sec:discussion}. The appendix
contains information on how to set up the Python software environment
for performing \gls{HPT} with \gls{SPOT} and \gls{tfruns}. Source code
for performing the experiments will included in the \gls{R} package
\gls{SPOT}. Further information are published on
\url{https://www.spotseven.de} and, with some delay, on CRAN
(\url{https://cran.r-project.org/package=SPOT}).

\section{Materials and Methods}\label{sec:materials}

\subsection{Hyperparameters}

Typical hyperparameters that are used to define \glspl{DNN}' are as
follows:

\begin{itemize}
\item optimization algorithms, e.g., \gls{RMSProp} (implemented in Keras as \RoptimizerRmsprop) or \gls{ADAM} (\RoptimizerAdam). These will be discussed in Sec.\ref{sec:optimizers}.
\item loss functions, e.g.,  \gls{MSE} (\RlossMeanSquaredError), \gls{MAE} (\RlossMeanAbsoluteError), or \gls{CCE} (\RlossCategoricalCrossentropy). 
The actual optimized objective is the mean of the output array across all datapoints.
\item learning rate
\item activation functions
\item number of hidden layers and hidden units
\item size of the training batches
\item weight initialization schemes
\item regularization penalties
\item dropout rates. Dropout is a commonly used regularization technique for \glspl{DNN}.
Applied to a layer, dropout consists of randomly setting to zero (dropping out) 
a percentage of output features of the layer during training \citep{chol18a}.
\item batch normalization
\end{itemize}

\begin{table}
\caption{Symbols used in this paper}
\label{tab:symbols}
\begin{tabulary}{\textwidth}{R R L  L }
\hline
Symbol  & &  Name & Comment, Example \\
\hline
$\lambda$  & &  hyperparameter configuration & \\
$\lambda_i$  & &  $i$-th hyperparameter configuration & used in \gls{SMBO} \\
$\lambda^{(*)}$  & &  best hyperparameter configuration & best configuration in theory\\
$\hatlambda$  & &  best hyperparameter configuration obtained by evaluating a finite set of samples& best configuration \enquote{in practice}\\
$\Lambda$  & &  hyperparameter space & \\
$G_x$ & & natural (ground truth) distribution & \\
$x$  & &  data point & \\
$X$  & &  data & usually partitioned into training, validation, and test data \\
$X^{(\text{train})} $  & &  training data & \\
$X^{(\text{valid})} $  & &  validation data & \\
$X^{(\text{test})} $  & &  test data & \\
$A$  & &  algorithm & \\
$t$  & &  iteration counter & counter for the \gls{SPOT} models, i.g., the $t$-th \gls{SPOT} metamodel will be denoted as $M(t)$\\
$\Psi$  & &  hyperparameter response space & \\
$\psi_i$  & &  hyperparameter response surface function evaluated for the $i$-th hyperparameter configuration $\lambda_i$\\
\psitrain  & &  hyperparameter response surface function (on train data) & \\
\psitest  & &  hyperparameter response surface function (on test data) & \\
\psival  & &  hyperparameter response surface function (on validation data) & \\
\hline
\end{tabulary}
\end{table}

\begin{example}[Conditionally dependent hyperparameters; \cite{Mend19a}]
This example illustrates that some hyperparameters are conditionally dependent 
on the number of layers. 
\cite{Mend19a} consider 
\begin{description}
\item[Network hyperparameters,] e.g., batch size, number of updates,
number of layers, learning rate, $L_2$ regularization, dropout output layer,
solver type (SGD, Momentum, \gls{ADAM}, Adadelta, Adagrad, smorm, Nesterov), 
learning-rate policy (fixed, inv, exp, step)
\item[Parameters conditioned on solver type,] e.g., $\beta_1$ and $\beta_2$, $\rho$, MOMENTUM,
\item[Parameters conditioned on learning-rate policy,] e.g., $\gamma$, $k$, and $s$,
\item[Per-layer hyperparameters,] e.g., activation-type (sigmoid, tanH, ScaledTanH, ELU, ReLU, Leaky, Linear), 
number of units, dropout in layer, weight initialization (Constant, Normal, Uniform, Glorot-Uniform, Glorot-Normal,
He-Normal), std. normal init., leakiness, tanh scale in/out.
\end{description}
For practical reasons, \cite{Mend19a} constrained the number of layers to be between one and six: 
firstly, they aimed to keep the training time of a single configuration low,
and secondly each layer adds eight per-layer hyperparameters to the configuration space, 
such that allowing additional layers would further complicate the configuration process.
\end{example}

\subsection{Hyperparameter: Features}

This section considers some properties, which are specific to \gls{DNN}
hyperparameters.

\subsubsection{Optimizers}\label{sec:optimizers}

\cite{Choi19a} considered \gls{RMSProp} with momentum \citep{Tiel12a},
\gls{ADAM} \citep{King16a} and \gls{ADAM} \citep{Doza16a} and claimed
that the following relations holds: \begin{equation*}
\begin{aligned}
    \sgd &\subseteq \momentum \subseteq \rmsprop\\
    \sgd &\subseteq \momentum \subseteq \adam\\
    \sgd &\subseteq \nesterov \subseteq \nadam
\end{aligned}
\end{equation*}

\begin{example}[ADAM can approximately simulate MOMENTUM]
MOMENTUM can be approximated with \gls{ADAM}, if a learning rate schedule that accounts for \gls{ADAM}’s bias correction is implemented. 
\end{example}

\cite{Choi19a} demonstrated that these inclusion relationships are
meaningful in practice. In the context of \gls{HPT} and \gls{HPO},
inclusion relations can significantly reduce the complexity of the
experimental design. These inclusion relations justify the selection of
a basic set, e.g., \gls{RMSProp}, \gls{ADAM}, and \gls{NADAM}.

\subsubsection{Batch Size}

\cite{Shal19a} and \cite{Zhan19a} have shown empirically that increasing
the batch size can increase the gaps between training times for
different optimizers.

\subsection{Performance Measures for Hyperparameter Tuning}
\subsubsection{Measures}

\citet{Kedz20a} state that \enquote{unsurprisingly},
\emph{accuracy}\footnote{Accuracy in binary classification is the proportion of correct predictions among the total number of observations \citep{Metz78a}.}
is considered as the most important performance measure. However, there
are many other ways to measure model quality, e.g., metrics based on
time complexity and robustness or the model complexity
(interpretability) \citep{bart20gArxiv}.

In contrast to classical optimization, where the same optimization
function can be used for tuning and final evaluation, training of
\glspl{DNN} faces a different situation:

\begin{itemize}
\item training is based on the loss function, 
\item whereas the final evaluation is based on a different measure, e.g., accuracy.
\end{itemize}

The loss function acts as a surrogate for the performance measure the
user is finally interested in. Several performance measures are used at
different stages of the \gls{HPO} procedures:

\begin{enumerate}
\item training loss, i.e., \psitrain,
\item training accuracy, i.e., \facctrain,
\item validation loss, i.e., \psival,
\item validation accuracy, i.e., \faccval,
\item test loss, i.e., \psitest, and 
\item test accuracy, i.e., \facctest.
\end{enumerate}

This complexity gives reason for the following question:

\begin{description}
\item[Question:] Which performance measure should be used during the \gls{HPT} (\gls{HPO}) procedure?
\end{description}

Most authors recommend using test accuracy or test loss as the measure
for hyperparameter tuning \citep{Schn19a}. In order to understand the
correct usage of these performance measures, it is important to look at
the goals, i.e., selection or assessment, of a tuning study.

\subsubsection{Model Selection and Assessment}

\cite{Hast17a} stated that selection and assessment are two separate
goals:

\begin{description}
\item[Model selection,] i.e., estimating the performance of different models in order
to choose the best one. Model selection is important \emph{during} the tuning procedure, whereas 
model assessment is used for the \emph{final} report (evaluation of the 
results).
\item[Model assessment,] i.e., having chosen a final model, estimating its prediction error (generalization error) on new data. 
Model assessment is performed to ascertain 
whether predicted values from the model are likely to accurately predict responses 
on future 
observations or samples not used to develop the model.
Overfitting is a major problem in this context.
\end{description}

In principle, there are two ways of model assessment and selection:
internal versus external. In the following, \(N\) denotes the total
number of samples.

\begin{description}
\item[External assessment] uses different sets of data. The first $m$ data samples are 
for model training and $N-m$ for validation. Problem: holding back data from model fitting results in lower precision and power.
\item[Internal Assessment] uses data splitting and resampling methods. The true error 
might be \emph{underestimated}, because the same data samples that were used for fitting the
model are used for prediction. The so-called in-sample (also apparent, or resubstitution) error
is smaller than the true error. 
\end{description}

In a data-rich situation, the best approach for both problems is to
randomly divide the dataset into three parts:

\begin{enumerate}
\item a training set to fit the models, 
\item a validation set to estimate prediction error for model selection, and 
\item a test set for assessment of the generalization error of the final chosen model. 
\end{enumerate}

The test set should be brought out only at the end of the data analysis.
It should not be used during the training and validation phase. If the
test set is used repeatedly, e.g., for choosing the model with smallest
test-set error,
\enquote{the test set error of the final chosen model will underestimate the true test error, sometimes substantially.}
\citep{Hast17a}

The following example \ref{ex:wils17a} shows that there is no general
agreement on how to use training, validation, and test sets as well as
the associated performance measures.

\begin{example}[Basic Comparisons in Manual Search]\label{ex:wils17a}
\cite{Wils17a} describe a manual search. 
They allocated a pre-specified budget on the number of epochs used for training each model. 
\begin{itemize}
\item When a test set was available, it was used to chose the settings that achieved the best 
peak performance on the test set by the end of the fixed epoch budget.
\item 
If no explicit test set was available, e.g., for \gls{CIFARTEN}, 
they chose the settings that achieved the lowest training loss at the end of the fixed epoch budget.
\end{itemize}
\end{example}

Theoretically, in-sample error is not usually of interest because future
values of the hyperparameters are not likely to coincide with their
training set values. \cite{Berg12a} stated that because of finite data
sets, test error is not monotone in validation error, and depending on
the set of particular hyperparameter values \(\lambda\) evaluated, the
test error of the best-validation error configuration may vary, e.g.,
\begin{equation}
\psitrain_i < \psitrain_j \centernot\implies \psitest_i < \psitest_j, 
\end{equation} where \(\psi_i^{(\cdot)}\) denotes the value of the
hyperparameter response surface for the \(i\)-th hyperparameter
configuration \(\lambda_i\).

Furthermore, the estimator, e.g., for loss, obtained by using a single
hold-out test set usually has high variance. Therefore, \gls{CV} methods
were proposed. \cite{Hast17a} concluded

\begin{quote}
\enquote{that estimation of test error for a particular training set is not easy in general, 
given just the data from that same training set. 
Instead, cross-validation and 
related methods may provide reasonable estimates of the expected error.}
\end{quote}

The standard practice for evaluating a model found by \gls{CV} is to
report the hyperparameter configuration that minimizes the loss on the
validation data, i.e., \(\hatlambda\) as defined in Equation
\ref{eq:hatlambda}. Repeated \gls{CV} is considered standard practice,
because it reduces the variance of the estimator. \(k\)-fold \gls{CV}
results in a more accurate estimate as well as in some information about
its distribution. There is, as always, a trade-off: the more \gls{CV}
folds the better the estimate, but more computational time is needed.

When different trials have nearly optimal validation means, then it is
not clear which test score to report: small changes in the
hyperparameter values could generate a different test error.

\begin{example}[Reporting the model assessment (final evaluation) \citep{Berg12a}]
When reporting performance of learning algorithms, it can be useful to take into 
account the uncertainty due to the choice of hyperparameters values. 
\cite{Berg12a} present a procedure for estimating test set accuracy, 
which takes into account any uncertainty in the choice of which trial is actually the best-performing one. 
To explain this procedure, they 
distinguish between estimates of performance 
$\psival$
and $\psitest$ 
based on the validation and test sets, respectively.

To resolve the difficulty of choosing a winner,
\cite{Berg12a} 
reported a weighted average of all the test set scores, 
in which each one is weighted by the probability that its particular 
$\lambda_s$ is in fact the best. 
In this view, the uncertainty arising from 
$X^{(\text{valid})}$
being a finite sample of the natural (ground truth) distribution $G_x$
makes the test-set score of the best model among 
$\{ \lambda_i \}_{i = 1,2, \ldots, S}$ a random variable, $z$.
\end{example}

\subsection{Practical Considerations}

Unfortunately, training, validation, and test data are used
inconsistently in \gls{HPO} studies: for example, \cite{Wils17a}
selected \emph{training loss}, \(\psitrain\), (and not validation loss)
during optimization and reported results on the test set \(\psitest\).

\cite{Choi19a} considered this combination as a
\enquote{somewhat non-standard choice} and performed tuning
(optimization) on the validation set, i.e., they used \psival for
tuning, and reported results \psitest on the test set. Their study
allows some valuable insight into the relationship of validation and
test error:

\begin{quote}
\enquote{For a \emph{relative comparison} between models during the tuning procedure,
in-sample error is convenient and often leads to effective model selection. 
The reason is that the relative (rather than absolute performance)
error is required for the comparisons.} \citep{Choi19a}
\end{quote}

\cite{Choi19a} compared the final predictive performance of \gls{NN}
optimizers after tuning the hyperparameters to minimize validation
error. They concluded that their
\enquote{final results hold regardless of whether they 
compare final validation error, i.e., \psival, or test error, i.e., \psitest}.
Figure 1 in \cite{Choi19a} illustrates that the relative performance of
optimizers stays the same, regardless of whether the validation or the
test error is used. \cite{Choi19a} considered two statistics: (i) the
quality of the best solution and (ii) the speed of training, i.e., the
number of steps required to reach a fixed validation target.

\subsubsection{Some Considerations about Cross Validation}

There are some drawbacks of \(k\)-fold \gls{CV}: at first, the choice of
the number of observations to be hold out from each fit is unclear: if
\(m\) denotes the size of the training set, with \(k = m\), the \gls{CV}
estimator is approximately unbiased for the true (expected) prediction
error, but can have high variance because the \(m\)
\enquote{training sets} are similar to one another. The computational
costs are relatively high, because \(m\) evaluations of the model are
necessary. Secondly, the number of repetitions needed to achieve
accurate estimates of accuracy can be large. Thirdly, \gls{CV} does not
fully represent variability of variable selection: if \(m\) subjects are
omitted each time from set of \(N\), the sets of variables selected from
each sample of size \(N-m\) are likely to be different from sets
obtained from independent samples of \(N\) subjects. Therefore, \gls{CV}
does not validate the full \(N\) subject model. Note, Monte-Carlo
\gls{CV} is an improvement over standard \gls{CV} \citep{Pica84a}.

\subsection{Related Work}

Before presenting the elements (benchmarks and software tools) for the
experiments in Sec. \ref{sec:setup}, we consider existing, related
approaches that might be worth looking at. This list is not complete and
will be updated in forthcoming versions of this paper.

\subsubsection{Hyperparameter Optimization Software and Benchmark Studies}\label{sec:studies}

\gls{SMBO} based on Kriging (aka Gaussian processes or \gls{BO}) has
been successfully applied to \gls{HPT} in several works, e.g.,
\cite{BaBM04} propose a combination of classical statistical tools,
\gls{BO} (\gls{DACE}), and \gls{CART} as a surrogate model. The
integration of \gls{CART} made \gls{SMBO} applicable to more general
\gls{HPT} problems, e.g., problems with categorical parameters.
\cite{Hutt11a} presented a similar approach by proposing \gls{SMAC} as a
tuner that is capable of handling categorical parameters by using
surrogate models based on random forests. Similar to the \gls{OCBA}
approach in \gls{SPOT}, \cite{Hutt11a} implemented an
\emph{intensification mechanism} for handling multiple instances. Early
\gls{SPOT} versions used a very simple intensification mechanism: (i)
the best solution is evaluated in each iteration and (ii) new candidate
solutions, that were proposed by the surrogate model, are evaluated as
often as the current best solution. This simple intensification strategy
was replaced by the more sophisticated \gls{OCBA} strategy in \gls{SPOT}
\citep{Bart11b}.

\gls{HPO} developed very quickly, new branches and extensions were
proposed, e.g., \gls{CASH}, \gls{NAS}, \gls{AutoHAS}, and further
\enquote{Auto-*} approaches \citep{Thor13a}, \citep{Dong20b}.
\cite{Kedz20a} analyzed what constitutes these systems and survey
developments in \gls{HPO}, multi-component models, \gls{NN} architecture
search, automated feature engineering, meta-learning, multi-level
ensembling, dynamic adaptation, multi-objective evaluation, resource
constraints, flexible user involvement, and the principles of
generalization. The authors developed a conceptual framework to
illustrate one possible way of fusing high-level mechanisms into an
autonomous \gls{ML} system. \emph{Autonomy} is considered as the
capability of \gls{ML} systems to independently adjust their results
even in dynamically changing environments. They discuss how \gls{AutoML}
can be transformed into \gls{AutonoML}, i.e, the systems are able to
independently \enquote{design, construct, deploy, and maintain} \gls{ML}
models similar to the \gls{CAAI} approach presented by \citet{Stro20a}.
Because \cite{Kedz20a} already presented a comprehensive overview of
this development, we will list the most relevant \enquote{highlights} in
the following.

\cite{Snoe12a} used the \gls{CIFARTEN} dataset, which consists of
\(60,000\) \(32 \times 32\) colour images in ten classes, for optimizing
the hyperparameters of a \glspl{CNN}.

\cite{Berg13a} proposed a meta-modeling approach to support automated
\gls{HPO}, with the goal of providing practical tools that replace
hand-tuning. They optimized a three layer \gls{CNN}.

\cite{Egge13a} collected a library of \gls{HPO} benchmarks and evaluated
three \gls{BO} methods. They considered the \gls{HPO} problem under
\(k\)-fold \gls{CV} as a minimization problem of \(\psival\) as defined
in Equation \ref{eq:psivalcv}.

\cite{Zoph17a} studied a new paradigm of designing \gls{CNN}
architectures and describe a scalable method to optimize these
architectures on a dataset of interest, for instance the ImageNet
classification dataset.

\cite{Bala18a} presented DeepHyper, a Python package that provides a
common interface for the implementation and study of scalable
hyperparameter search methods.

\cite{Karm18a} created a \enquote{\emph{Rosetta Stone}} of \gls{DL}
frameworks to allow data-scientists to easily leverage their expertise
from one framework to another. They provided a common setup for
comparisons across GPUs (potentially CUDA versions and precision) and
for comparisons across languages (Python, Julia, R). Users should be
able to verify expected performance of own installation.

\cite{Mazz19b} introduced a \gls{NAS} framework to improve keyword
spotting and spoken language identification models.

\cite{Mend19a} introduced Auto-Net, a system that automatically
configures \gls{NN} with \gls{SMAC} by following the same AutoML
approach as Auto-WEKA and Auto-sklearn. They achieved the best
performance on two datasets in the human expert track of an
AutoMLChallenge.

\cite{omal19a} presented \emph{Keras tuner}, a hyperparameter tuner for
Keras with TensorFlow 2.0. They defined a model-building function, which
takes an argument from which hyperparameters such as the units (hidden
nodes) of the neural network. Available tuners are RandomSearch and
Hyperband.

Because optimizers can affect the \gls{DNN} performance significantly,
several tuning studies devoted to optimizers were published during the
last years: \cite{Schn19a} introduced a benchmarking framework called
\gls{DeepOBS}, which includes a wide range of realistic \gls{DL}
problems together with standardized procedures for evaluating
optimizers. \cite{Schm21a} performed an extensive, standardized
benchmark of fifteen particularly popular \gls{DL} optimizers.

A highly recommended study was performed by \cite{Choi19a}, who
presented a taxonomy of first-order optimization methods. Furthermore,
\cite{Choi19a} demonstrated the sensitivity of optimizer comparisons to
the hyperparameter tuning protocol. Optimizer rankings can be changed
easily by modifying the hyperparameter tuning protocol. Their findings
raised serious questions about the practical relevance of conclusions
drawn from certain ways of empirical comparisons. They also claimed that
tuning protocols often differ between works studying \gls{NN} optimizers
and works concerned with training \glspl{NN} to solve specific problems.

\cite{Zimm20a} developed Auto-PyTorch, a framework for \gls{AutoDL} that
uses \gls{BOHB} as a backend to optimize the full \gls{DL} pipeline,
including data preprocessing, network training techniques and
regularization methods.

\cite{Mazz21a} presented Google's Model Search, which is an open source
platform for finding optimal \gls{ML} models based on \gls{TF}. It does
not focus on a specific domain.

\citet{Wist19a} described how complex \gls{DL} architectures can be seen
as combinations of a few elements, so-called \emph{cells}, that are
repeated to build the complete network. \citet{Zoph17b} were the first
who proposed a cell-based approach, i.e., choices made about a \gls{NN}
architecture is the set of meta-operations and their arrangement within
the cell. Another interesting example are function-preserving morphisms
implemented by the Auto-Keras package to effectively traverse potential
networks \citet{Jin19b}.

Tunability is an interesting concept that should be mentioned in the
context of \gls{HPT} \citep{Prob19a}. The term \emph{tunability}/
describes a measure for modeling algorithms as well as for individual
hyperparameters. It is the difference between the model quality for
default values (or reference values) and the model quality for optimized
values (after \gls{HPT} is completed). Or in the words of
\citet{Prob19a}:
\enquote{measures for quantifying the tunability of the whole algorithm
and specific hyperparameters based on the differences between the performance of default
hyperparameters and the performance of the hyperparameters when this hyperparameter is
set to an optimal value}. Tunability of individual hyperparameters can
also be used as a measure of their \emph{relevance}, \emph{importance},
or \emph{sensitivity}. Hyperparameters with high tunability are
accordingly of greater importance for the model. The model reacts
strongly to (i.e., is sensitive to) changes in these hyperparameters.
The hope is that identifying \emph{tunable} hyperparameters, i.e., ones
that model performance is particularly sensitive to, will allow other
settings to be ignored, constraining search space. Unfortunately,
tunability strongly depends on the choice of the dataset, which makes
generalization of results very difficult.

To conclude this overview, we would like to mention relevant criticism
of \gls{HPO}: some publications even claimed that extensive \gls{HPO} is
not necessary.

\begin{itemize}
\item 
\cite{Eric20a} introduced a framework (AutoGluon-Tabular) that \enquote{requires only a single line of Python to train highly accurate machine learning models on an unprocessed tabular dataset such as a CSV file}. 
AutoGluon-Tabular ensembles several models and stacks them in multiple layers. 
The authors claim that AutoGluon-Tabular outperforms \gls{AutoML} platforms such as 
TPOT, H2O, AutoWEKA, auto-sklearn, AutoGluon, and Google AutoML Tables. 
\item 
\citet{Yu20a} claimed that the evaluated state-of-the-art \gls{NAS} algorithms do not surpass random search by a significant margin, 
and even perform worse in the \gls{RNN} search space.
\citet{Bala18b} reported a multitude of issues when attempting to execute automatic \gls{ML} frameworks. For example, regarding the random process, the authors state that \enquote{one common failure is in large multi-class classification tasks in which one of the classes lies entirely on one side of the train test split}. 
\item \citet{Liu18a} remarks that \enquote{for most existent AutoML works, regardless of the number of layers of the outer-loop algorithms, the configuration of the outermost layer is definitely done by human experts}. Human experts are shifted to a higher level, and are still in the loop.
\item \citet{Li19a} stated that (i) better baselines that accurately quantify the performance gains of \gls{NAS} methods, (ii) ablation studies (to learn about the \gls{NN} by removing parts of it) that isolate the impact of individual \gls{NAS} components, and (iii) reproducible results that engender confidence and foster scientific progress are necessary.
\end{itemize}

\subsubsection{Artificial Toy Functions}

Because \gls{BO} does not work well on high-dimensional mixed continuous
and categorical configuration spaces, \cite{Falk18a} used a simple
counting ones problem to analyze this problem. \cite{Zaef16b} discussed
these problems in greater detail. How to implement \gls{BO} for discrete
(and continuous) optimization problems was analyzed in the seminal paper
by \cite{Bart16n}.

\subsubsection{Experiments on Surrogate Benchmarks}

\cite{Falk18a} optimized six hyperparameters that control the training
procedure of a fully connected \gls{DNN} (initial learning rate, batch
size, dropout, exponential decay factor for learning rate) and the
architecture (number of layers, units per layer) for six different
datasets gathered from OpenML \citep{vans14a}, see Table
\ref{tab:falk18a}.

\begin{table}[tb]
\centering
\caption{The hyperparameters and architecture choices for the fully connected networks as defined in \cite{Falk18a}}
\label{tab:falk18a}
\begin{tabular}{lrrr}
\hline
    Hyperparameter & Lower Bound& Upper Bound & Log-transform\\
    \hline
    batch size & $2^3$& $2^8$  & yes\\
    dropout rate & $0$ & $0.5$ & no\\
    initial learning rate & $1e-6$ & $1e-2$ & yes\\
    exponential decay factor & $-0.185$ &  $0$ & no\\
    \# hidden layers & $1$ & $5$ & no\\
    \# units per layer & $2^{4}$ & $2^{8}$ & yes\\
\hline
\end{tabular}
\end{table}

\cite{Falk18a} used a surrogate \gls{DNN} as a substitute for training
the networks directly, . To build a surrogate, they sampled \(10,000\)
random configurations for each data set, trained them for \(50\) epochs,
and recorded classification error after each epoch, and total training
time. Two independent random forests models were fitted to predict these
two quantities as a function of the hyperparameter configuration used.
\cite{Falk18a} noted that \gls{HB} initially performed much better than
the vanilla \gls{BO} methods and achieved a roughly three-fold speedup
over random search.

Instead of using a surrogate network, we will use the original
\glspl{DNN}. Our approach is described in Sec.\ref{sec:experiments}.

\subsection{Stochasticity}

Results from \gls{DNN} tuning runs are noisy, e.g., caused by random
sampling of batches and initial parameters. Repeats to estimate means
and variances that are necessary for a sound statistical analysis
require substantial computational costs.

\subsection{Software: Keras, Tensorflow, tfruns, and SPOT }\label{sec:setup}
\subsubsection{Keras and Tensorflow}

Keras is the high-level \gls{API} of \gls{TF}, which is developed with a
focus on enabling fast experimentation. \gls{TF} is an open source
software library for numerical computation using data flow graphs
\citep{abad16a}. Nodes in the graph represent mathematical operations,
while the graph edges represent the multidimensional data arrays
(tensors) communicated between them \citep{omal19a}. The
\texttt{tensorflow} \gls{R} package provides access to the complete
\gls{TF} \gls{API} from within \gls{R}.

\subsubsection{The R Package tfruns}

The \gls{R} package
\gls{tfruns}\footnote{\url{https://cran.r-project.org/package=tfruns}, \url{https://tensorflow.rstudio.com/tools/tfruns}}
provides a suite of tools for tracking, visualizing, and managing
\gls{TF} training runs and experiments from \gls{R}. \gls{tfruns}
enables tracking the hyperparameters, metrics, output, and source code
of every training run and comparing hyperparameters and metrics across
runs to find the best performing model. It automatically generates
reports to visualize individual training runs or comparisons between
runs. \gls{tfruns} can be used without any changes to source code,
because run data is automatically captured for all Keras and \gls{TF}
models.

\subsubsection{SPOT}

The \gls{SPOT} package for \gls{R} is a toolbox for tuning and
understanding simulation and optimization algorithms \citep{bart21b}.
\gls{SMBO} investigations are common approaches in simulation and
optimization. Sequential parameter optimization has been developed,
because there is a strong need for sound statistical analysis of
simulation and optimization algorithms. \gls{SPOT} includes methods for
tuning based on classical regression and analysis of variance
techniques; tree-based models such as \gls{CART} and random forest;
\gls{BO} (Gaussian process models, aka Kriging), and combinations of
different meta-modeling approaches.

\gls{SPOT} implements key techniques such as exploratory fitness
landscape analysis and sensitivity analysis. \gls{SPOT} can be used for
understanding the performance of algorithms and gaining insight into
algorithm's behavior. Furthermore, \gls{SPOT} can be used as an
optimizer and for automatic and interactive tuning. \gls{SPOT} finds
improved solutions in the following way:

\begin{enumerate}
\item
  Initially, a population of (random) solutions is created.
\item
  A set of surrogate models is specified.
\item
  Then, the solutions are evaluated on the objective function.
\item
  Next, surrogate models are built.
\item
  A global search is performed on the surrogate model(s) to generate new candidate solutions.
\item
  The new solutions are evaluated on the objective function, e.g., the loss is determined.
\end{enumerate}

These steps are repeated, until a satisfying solution has been found as
described in \cite{bart21b}.

\paragraph{SPOT Surrogate Models}

\gls{SPOT} performs model selection during the tuning run: training data
\(\Xtrain\) is used for fitting (training) the models, e.g., the weights
of the \glspl{DNN}. Each trained model
\(A_{\lambda_i}\left( \Xtrain \right)\) will be evaluated on the
validation data \(\Xval\), i.e., the loss is calculated as
\begin{equation}
\psival_i  = \frac{1}{| \Xval|} \sum_{x \in \Xval} L \left( x; A_{\lambda_i} (\Xtrain) \right).
\end{equation} Based on \((\lambda_i, \psival_i )\), a surrogate model
\(M(t)\) is fitted, e.g., a \gls{BO} (Kriging) model using \gls{SPOT}'s
\RbuildKriging function. For each hyperparameter configuration
\(\lambda_i\), \gls{SPOT} reports information about the related
\gls{DNN} models \(A_{\lambda_i}\)

\begin{enumerate}
\item training loss, $\psitrain$,
\item training accuracy, $\facctrain$, 
\item validation (testing) loss, $\psival$,  and 
\item validation (testing) accuracy, $\faccval$.
\end{enumerate}

Output from a typical run is show in Figure \ref{fig:trainResult}.

\section{Experiments: Tuning Hyperparameters with SPOT}\label{sec:experiments}

How the software packages (Keras, \gls{TF}, \gls{tfruns}, and
\gls{SPOT}) can be combined in a very efficient and effective manner
will be exemplified in this section. The general \gls{DNN} workflow is
as follows: first the training data, \texttt{train\_images} and
\texttt{train\_labels} are fed to the \gls{DNN}. The \gls{DNN} will then
learn to associate images and labels. Based on the Keras parameter
\texttt{validation\_split}, the training data will be partitioned into a
(smaller) training data set and a validation data set. The corresponding
code is shown in Sec. \ref{sec:mnistex}. The trained \gls{DNN} produces
predictions for validations.

\subsection{The Data Set: MNIST}\label{sec:mnistex}

The \gls{DNN} in this example uses the Keras \gls{R} package to learn to
classify hand-written digits from the \gls{MNIST} data set. This is a
supervised multi-class classification problem, i.e., grayscale images of
handwritten digits (\(28 \times 28\) pixels) should be assigned to ten
categories (0 to 9). \gls{MNIST} is a set of \(60,000\) training and
\(10,000\) test images. The \gls{MNIST} data set is included in Keras as
\texttt{train} and \texttt{test} lists, each of which includes a set of
images (\texttt{x}) and associated labels (\texttt{y}):
\texttt{train\_images} and \texttt{train\_labels} form the training set,
the data that the \gls{DNN} will learn from. The \gls{DNN} can be tested
on the \Xtest set (\texttt{test\_images} and \texttt{test\_labels}). The
images are encoded as as 3D arrays, and the labels are a 1D array of
digits, ranging from 0 to 9.

Before training the \gls{DNN}, the data are preprocessed by reshaping it
into the shape the \gls{DNN} can process. The natural (original)
training images were stored in an array of shape \((60000, 28, 28)\) of
type \texttt{integer} with values in the \([0, 255]\) interval. They are
transformed into a double array of shape \((60000, 28 \times 28)\) with
\gls{RGB} values between \(0\) and \(1\), i.e., all values will be
scaled that they are in the \([0, 1]\) interval. Furthermore, the labels
are categorically encoded.

\begin{Shaded}
\begin{Highlighting}[]
\NormalTok{mnist }\OtherTok{\textless{}{-}} \FunctionTok{dataset\_mnist}\NormalTok{()}
\NormalTok{x\_train }\OtherTok{\textless{}{-}}\NormalTok{ mnist}\SpecialCharTok{$}\NormalTok{train}\SpecialCharTok{$}\NormalTok{x}
\NormalTok{y\_train }\OtherTok{\textless{}{-}}\NormalTok{ mnist}\SpecialCharTok{$}\NormalTok{train}\SpecialCharTok{$}\NormalTok{y}
\NormalTok{x\_test }\OtherTok{\textless{}{-}}\NormalTok{ mnist}\SpecialCharTok{$}\NormalTok{test}\SpecialCharTok{$}\NormalTok{x}
\NormalTok{y\_test }\OtherTok{\textless{}{-}}\NormalTok{ mnist}\SpecialCharTok{$}\NormalTok{test}\SpecialCharTok{$}\NormalTok{y}

\FunctionTok{dim}\NormalTok{(x\_train) }\OtherTok{\textless{}{-}} \FunctionTok{c}\NormalTok{(}\FunctionTok{nrow}\NormalTok{(x\_train), }\DecValTok{784}\NormalTok{)}
\FunctionTok{dim}\NormalTok{(x\_test) }\OtherTok{\textless{}{-}} \FunctionTok{c}\NormalTok{(}\FunctionTok{nrow}\NormalTok{(x\_test), }\DecValTok{784}\NormalTok{)}

\NormalTok{x\_train }\OtherTok{\textless{}{-}}\NormalTok{ x\_train }\SpecialCharTok{/} \DecValTok{255}
\NormalTok{x\_test }\OtherTok{\textless{}{-}}\NormalTok{ x\_test }\SpecialCharTok{/} \DecValTok{255}

\NormalTok{y\_train }\OtherTok{\textless{}{-}} \FunctionTok{to\_categorical}\NormalTok{(y\_train, }\DecValTok{10}\NormalTok{)}
\NormalTok{y\_test }\OtherTok{\textless{}{-}} \FunctionTok{to\_categorical}\NormalTok{(y\_test, }\DecValTok{10}\NormalTok{)}
\end{Highlighting}
\end{Shaded}

\subsection{The Neural Network}\label{sec:nn}

The \gls{DNN} consists of a sequence of two dense (fully connected)
layers. The second layer is a ten-way \texttt{softmax} layer that
returns an array of ten probability scores. Each score represents the
probability that the input image belongs to one of the ten \gls{MNIST}
digit classes.

\begin{Shaded}
\begin{Highlighting}[]
\NormalTok{model }\OtherTok{\textless{}{-}} \FunctionTok{keras\_model\_sequential}\NormalTok{()}
\NormalTok{model }\SpecialCharTok{\%\textgreater{}\%}
    \FunctionTok{layer\_dense}\NormalTok{(}\AttributeTok{units =} \DecValTok{256}\NormalTok{, }\AttributeTok{activation =} \StringTok{\textquotesingle{}relu\textquotesingle{}}\NormalTok{, }\AttributeTok{input\_shape =} \FunctionTok{c}\NormalTok{(}\DecValTok{784}\NormalTok{)) }\SpecialCharTok{\%\textgreater{}\%}
    \FunctionTok{layer\_dropout}\NormalTok{(}\AttributeTok{rate =} \FloatTok{0.4}\NormalTok{) }\SpecialCharTok{\%\textgreater{}\%}
    \FunctionTok{layer\_dense}\NormalTok{(}\AttributeTok{units =} \DecValTok{128}\NormalTok{, }\AttributeTok{activation =} \StringTok{\textquotesingle{}relu\textquotesingle{}}\NormalTok{) }\SpecialCharTok{\%\textgreater{}\%}
    \FunctionTok{layer\_dropout}\NormalTok{(}\AttributeTok{rate =} \FloatTok{0.3}\NormalTok{) }\SpecialCharTok{\%\textgreater{}\%}
    \FunctionTok{layer\_dense}\NormalTok{(}\AttributeTok{units =} \DecValTok{10}\NormalTok{, }\AttributeTok{activation =} \StringTok{\textquotesingle{}softmax\textquotesingle{}}\NormalTok{)}
\end{Highlighting}
\end{Shaded}

Finally, (i) the \emph{loss function}, which determines how the
\gls{DNN} good a prediction is based on the training data, (ii) the
\emph{optimizer}, i.e., the update mechanism of the network, which
adjusts the weights using backpropagation, and (iii) the \emph{metrics},
which monitor the progress during training and testing, are specified
using the \texttt{compile} function from Keras.

\begin{Shaded}
\begin{Highlighting}[]
\NormalTok{model }\SpecialCharTok{\%\textgreater{}\%} \FunctionTok{compile}\NormalTok{(}
    \AttributeTok{loss =} \StringTok{\textquotesingle{}categorical\_crossentropy\textquotesingle{}}\NormalTok{,}
    \AttributeTok{optimizer =} \FunctionTok{optimizer\_rmsprop}\NormalTok{(}\AttributeTok{lr =} \FloatTok{0.01}\NormalTok{),}
    \AttributeTok{metrics =} \FunctionTok{c}\NormalTok{(}\StringTok{\textquotesingle{}accuracy\textquotesingle{}}\NormalTok{)}
\NormalTok{)}
\end{Highlighting}
\end{Shaded}

The \gls{DNN} training can be started as follows (using Kera's
\texttt{fit} function).

\begin{Shaded}
\begin{Highlighting}[]
\NormalTok{history }\OtherTok{\textless{}{-}}\NormalTok{ model }\SpecialCharTok{\%\textgreater{}\%}\NormalTok{ keras}\SpecialCharTok{::}\FunctionTok{fit}\NormalTok{(}
\NormalTok{    x\_train, y\_train,}
    \AttributeTok{batch\_size =} \DecValTok{128}\NormalTok{,}
    \AttributeTok{epochs =} \DecValTok{20}\NormalTok{,}
    \AttributeTok{verbose =} \DecValTok{0}\NormalTok{,}
    \AttributeTok{validation\_split =} \FloatTok{0.2}
\NormalTok{)}
\end{Highlighting}
\end{Shaded}

\begin{figure}
\centering
\includegraphics{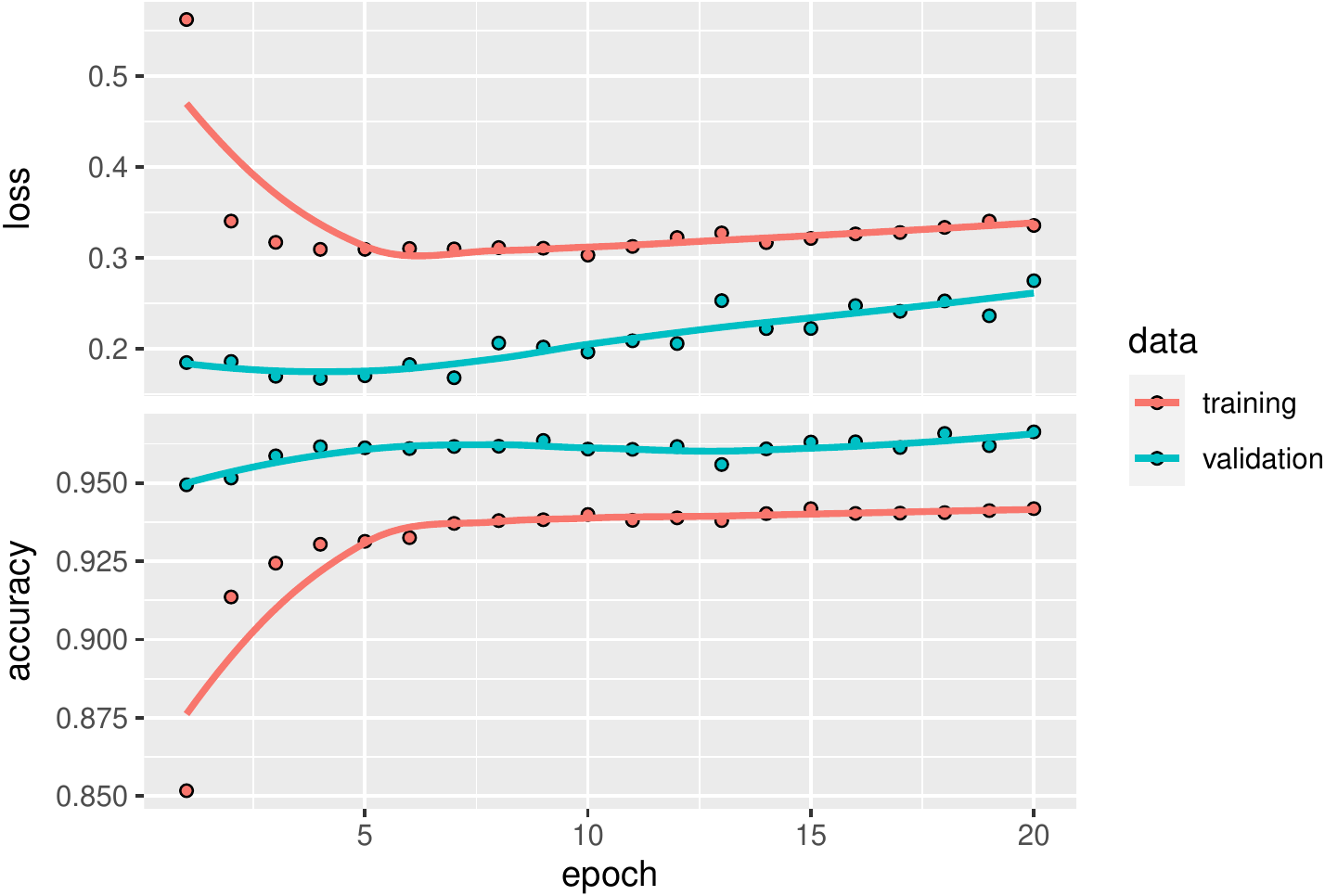}
\caption{\label{fig:trainResult}Training and validation data. Loss and
accuracy plotted against epochs.}
\end{figure}

Figure \ref{fig:trainResult} shows the quantities that are being
displayed during training:

\begin{enumerate}[(i)]
\item  the \emph{loss} of the network over the training and validation data, 
$\psitrain$ and $\psival$, respectively,
and 
\item the \emph{accuracy} of the network over the training and validation data, $\facctrain$ and $\faccval$, respectively. 
\end{enumerate}

This figure illustrates that an accuracy greater than 95 percent on the
training data can be reached quickly.

Finally, using Kera's \texttt{evaluate} function, the \gls{DNN} model
performance can be checked on \Xtest.

\begin{Shaded}
\begin{Highlighting}[]
\NormalTok{score }\OtherTok{\textless{}{-}}\NormalTok{ model }\SpecialCharTok{\%\textgreater{}\%} \FunctionTok{evaluate}\NormalTok{(}
\NormalTok{    x\_test, y\_test,}
    \AttributeTok{verbose =} \DecValTok{0}
\NormalTok{)}
\FunctionTok{cat}\NormalTok{(}\StringTok{\textquotesingle{}Test loss:\textquotesingle{}}\NormalTok{, score[[}\DecValTok{1}\NormalTok{]], }\StringTok{\textquotesingle{}}\SpecialCharTok{\textbackslash{}n}\StringTok{\textquotesingle{}}\NormalTok{)}
\FunctionTok{cat}\NormalTok{(}\StringTok{\textquotesingle{}Test accuracy:\textquotesingle{}}\NormalTok{, score[[}\DecValTok{2}\NormalTok{]], }\StringTok{\textquotesingle{}}\SpecialCharTok{\textbackslash{}n}\StringTok{\textquotesingle{}}\NormalTok{)}
\end{Highlighting}
\end{Shaded}

\begin{verbatim}
Test loss: 0.2828377 
Test accuracy: 0.9638 
\end{verbatim}

The relationship between \psitrain, \psival, and \psitest as well as
between \facctrain, \faccval, and \facctest can be analyzed with
\gls{SPOT}.

Running the \gls{DNN} model as a standalone process \emph{before}
starting the tuning process is strongly recommended. As shown in this
section, the default \gls{DNN} model seems to work fine.

\subsection{Interfacing tfruns from SPOT}

After testing the model as a standalone implementation, the model can be
combined with the \gls{SPOT} framework. A wrapper function is used to
connect \gls{tfruns} to the \gls{SPOT} tuner. The setup requires a few
lines of \gls{R} code only. Instead of two hyperparameters,
\texttt{var1} and \texttt{var2}, that are passed to \gls{TF} as shown in
the following code example, an arbitrary amount of hyperparameters can
be passed.

\begin{Shaded}
\begin{Highlighting}[]
\NormalTok{funTfrunsSingle }\OtherTok{\textless{}{-}} \ControlFlowTok{function}\NormalTok{(x) \{}
\NormalTok{    runs }\OtherTok{\textless{}{-}} \FunctionTok{tuning\_run}\NormalTok{(}\StringTok{"kerasModel.R"}\NormalTok{,}
       \AttributeTok{flags =} \FunctionTok{list}\NormalTok{(}\AttributeTok{var1 =}\NormalTok{ x[}\DecValTok{1}\NormalTok{], }\AttributeTok{var2 =}\NormalTok{ x[}\DecValTok{2}\NormalTok{]), }
       \AttributeTok{confirm =} \ConstantTok{FALSE}\NormalTok{)}
\NormalTok{    runs}\SpecialCharTok{$}\NormalTok{metric\_val\_loss[[}\DecValTok{1}\NormalTok{]]}
\NormalTok{\}}
\NormalTok{funTfruns }\OtherTok{\textless{}{-}} \FunctionTok{wrapFunction}\NormalTok{(funTfrunsSingle)}
\end{Highlighting}
\end{Shaded}

The first line defines the \gls{R} function \texttt{funTfrunsSingle()}
for a single hyperparameter configuration. It calls the code from the
\texttt{kerasModel.R} file, which implements the \gls{DNN} described in
Sec. \ref{sec:mnistex}.

In order to evaluate several hyperparameter configurations during one
single function call, \gls{SPOT}'s \texttt{wrapFunction()} is applied to
the \texttt{funTfrunsSingle()} function. Note, that \gls{SPOT} operates
on \texttt{matrix} objects.

\subsection{Hyperparameter Tuning with SPOT}

The following hyperparameters will be tuned:

\begin{enumerate}
\item[$x_1$, $x_2$]  the \emph{dropout rates}.
The dropout rates of the first and second layer will be tuned individually.
\item[$x_3$, $x_4$] the \emph{number of units}, i.e., the number of single outputs from a single layer. 
The number of units of the first and second layer will be tuned individually.
\item[$x_5$] the \emph{learning rate}, which controls how much to change the \gls{DNN} model in response 
to the estimated error each time the model weights are updated.
\item[$x_6$] the \emph{number of training epochs}, where a training epoch is one forward and backward pass of a 
complete data set.
\item[$x_7$] the \emph{batch size}, and
\item[$x_8$] \RoptimizerRmsprop's \emph{decay factor}.
\end{enumerate}

\begin{table}[tb]
\centering
\caption{The hyperparameters and architecture choices for the first \gls{DNN} example: fully connected networks}
\label{tab:hyper1}
\begin{tabular}{lllrrr}
\hline
    Variable Name& Hyperparameter & Type & Default  & Lower Bound & Upper Bound \\
    \hline
$x_1$ & first layer dropout rate & numeric & $0.4$  & $1e-6$ & $1$  \\
$x_2$ & second layer dropout rate & numeric & $0.3$ & $1e-6$ & $1$ \\
$x_3$ & units per first layer & integer  & $256$ & $16$ &  $512$ \\
$x_4$ & units per second layer & integer & $128$  & $4$ &  $256$ \\
$x_5$ & learning rate & numeric & $0.001$ & $0.0001$ & $0.1$ \\
$x_6$ & training epochs & integer & $20$  &$5$ & $25$ \\
$x_7$ & batch size & integer & $64$ & $8$ & $256$ \\
$x_8$ & rho & numeric & $0.9$ & $0.5$ & $0.999$ \\
\hline
\end{tabular}
\end{table}

These hyperparameters and their ranges are listed in Table
\ref{tab:hyper1}. Using these parameter specifications, we are ready to
perform the first \gls{SPOT} \gls{HPT} run:

\begin{Shaded}
\begin{Highlighting}[]
\NormalTok{res }\OtherTok{\textless{}{-}} \FunctionTok{spot}\NormalTok{(}
    \AttributeTok{x =} \ConstantTok{NULL}\NormalTok{,}
    \AttributeTok{fun =}\NormalTok{ funTfruns,}
\NormalTok{  lower }\OtherTok{\textless{}{-}} \FunctionTok{c}\NormalTok{(}\FloatTok{1e{-}6}\NormalTok{, }\FloatTok{1e{-}6}\NormalTok{, }\DecValTok{16}\NormalTok{, }\DecValTok{16}\NormalTok{, }\FloatTok{1e{-}9}\NormalTok{, }\DecValTok{10}\NormalTok{, }\DecValTok{16}\NormalTok{, }\FloatTok{0.5}\NormalTok{),}
\NormalTok{  upper }\OtherTok{\textless{}{-}} \FunctionTok{c}\NormalTok{(}\FloatTok{0.5}\NormalTok{, }\FloatTok{0.5}\NormalTok{, }\DecValTok{512}\NormalTok{, }\DecValTok{256}\NormalTok{, }\FloatTok{1e{-}2}\NormalTok{, }\DecValTok{50}\NormalTok{, }\DecValTok{512}\NormalTok{, }\DecValTok{1}\FloatTok{{-}1e{-}3}\NormalTok{),}
    \AttributeTok{control =} \FunctionTok{list}\NormalTok{(}
        \AttributeTok{funEvals =} \DecValTok{480}\NormalTok{,}
        \AttributeTok{types =} \FunctionTok{c}\NormalTok{(}
            \StringTok{"numeric"}\NormalTok{,}
            \StringTok{"numeric"}\NormalTok{,}
            \StringTok{"integer"}\NormalTok{,}
            \StringTok{"integer"}\NormalTok{,}
            \StringTok{"numeric"}\NormalTok{,}
            \StringTok{"integer"}\NormalTok{,}
            \StringTok{"integer"}\NormalTok{,}
            \StringTok{"numeric"}
\NormalTok{        )}
\NormalTok{    )}
\NormalTok{)}
\end{Highlighting}
\end{Shaded}

The budget is set to \(n=480\) evaluations (20 times the number of
hyperparameters (8), multiplied by the number of repeats (3)), i.e., the
total number of \gls{DNN} training and testing iterations.

\gls{SPOT} provides several options for adjusting the \gls{HPT}
parameters, e.g., type of the \gls{SMBO} model and optimizer as well as
the size of the initial design. These parameters can be passed via the
\texttt{spotControl} function to \texttt{SPOT}. For example, instead of
the default model, which is \gls{BO}, a random forest can be chosen. A
detailed description of the \gls{SPOT} tuning algorithm can be found in
\cite{bart21b}.

\section{Results}\label{sec:results}

While discussing the hyperparameter tuning results, \gls{HPT} does not
look the the final, best solution only. For sure, the hyperparameter
practitioner is interested in the best solution. But even from this
\emph{greedy} point of view, considering the
\emph{route to the solution} can is also of great importance, because
analysing this route enables \emph{learning} and can be much more
efficient in the long run compared to a greedy strategy.

\begin{example}
Consider a classification task that 
has to be performed several times in a different context with
similar data.
Instead of blindly (automatically) running the \gls{HPO} procedure individually for each classification task (which might also require a significant amount of time and resources, even when it is performed automatically) a few \gls{HPT} procedures are performed.
Insights gained from \gls{HPT} might help to avoid pitfalls such as
ill specified parameter ranges, too short run times, etc.
\end{example}

In addition to an effective and efficient way to determine the optimal
hyperparameters, \gls{SPOT} provides tools for
\emph{learning and understanding}.\footnote{Or, as \cite{Bart06a} wrote: `` [\gls{SPOT}] provides means for understanding algorithms’ performance (we will use datascopes similar to microscopes in biology and telescopes in astronomy).''}

The \gls{HPT} experiment from Sec. \ref{sec:experiments} used \(n=480\)
\gls{DNN} evaluations, i.e., \gls{SPOT} generated a \emph{result list}
(\texttt{res}) with the information shown in Fig.\ref{fig:res}.

\begin{figure}
\caption{Internal `list` structure of the result object `res`  from the \gls{SPOT} run.
\label{fig:res}
}
\begin{verbatim}
List of 9
 $ xbest   : num [1, 1:8] 4.67e-01 1.48e-01 4.31e+02 9.20e+01 1.97e-04 ...
 $ ybest   : num [1, 1] 0.0685
 $ x       : num [1:480, 1:8] 0.433 0.134 0.339 0.222 0.215 ...
 $ y       : num [1:480, 1] 0.8908 0.0837 0.1272 0.1418 0.1203 ...
 $ logInfo : logi NA
 $ count   : int 480
 $ msg     : chr "budget exhausted"
 $ modelFit:List of 33
  ..$ thetaLower      : num 1e-04
  ..$ thetaUpper      : num 100
  ..$ types           : chr [1:8] "numeric" "numeric" "integer" "integer" ...
  ...
  ..$ min             : num 0.0717
  ..- attr(*, "class")= chr "kriging"
 $ ybestVec: num [1:280] 0.0824 0.0824 0.0824 0.0824 0.0824 0.0824 0.0824 0.0824 0.0824 0.0824 ...
 \end{verbatim}
 \end{figure}

\begin{figure}
\centering
\includegraphics{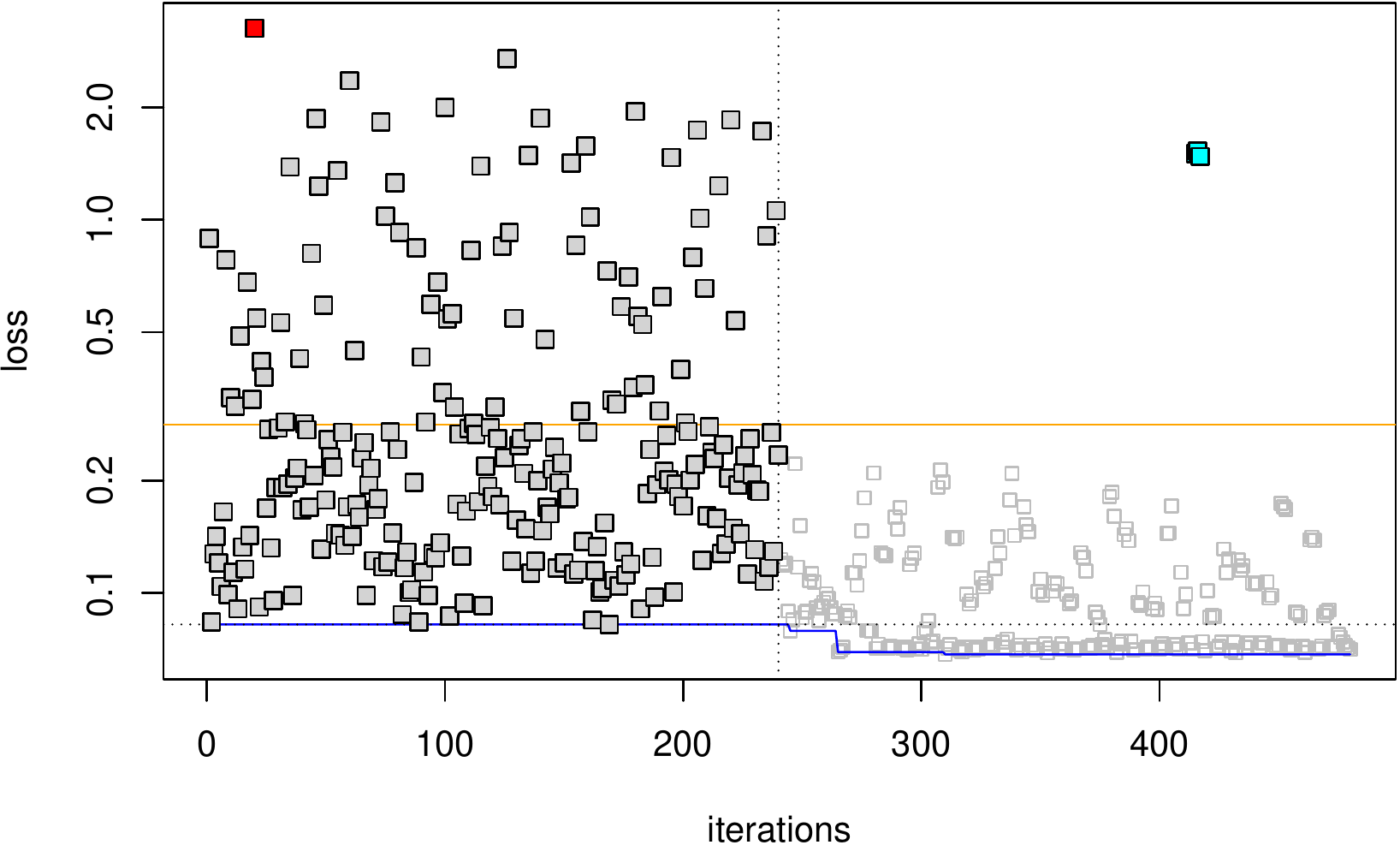}
\caption{\label{fig:resValues}Loss function values plotted against the
number of iterations. The \texttt{orange} line represents the loss
obtained with the default DNN hyperparameters. The dotted \texttt{black}
line represents the best loss value from the initial design. Initial
design points have black boxes. The \texttt{blue} line shows the best
function value found during the tuning procedure. \texttt{Grey} squares
represent the values generated during the hyperparameter tuning
procedure. The \texttt{red} square shows one large value, and
\texttt{cyan} colored dots indicate worse configurations that occurred
during the tuning procedure. These values should be investigated
further. Note: loss values plotted on a log scale}
\end{figure}

\emph{Plots.} First of all, the \texttt{res} list information will be
used to visualize the \emph{route to the solution}: in Fig.
\ref{fig:resValues}, loss function values are plotted against the number
of iterations. This figure reveals that some hyperparameter
configurations should be investigated further, because these
configurations result in relatively high loss function values. Using the
default hyperparameter configuration results in a loss value of
\(0.28\). The related hyperparameters values are shown in Table
\ref{tab:yLarge}.

\begin{table}[ht]
\caption{Worse configurations}
\label{tab:yLarge}
\centering
\begin{tabular}{rrrrrrrrr}
  \hline
& dropout1 & dropout2 & units0 & units1 & lr & epochs & batchSize & rho \\ 
 \hline
red & 0.03 & 0.42 & 133.00 & 198.00 & 0.01 & 48.00 & 43.00 & 0.58 \\ 
cyan & 0.05 & 0.05 & 295.00 & 163.00 & 0.00 & 10.00 & 449.00 & 0.51 \\ 
   \hline
\end{tabular}
\end{table}

\begin{figure}
\centering
\includegraphics{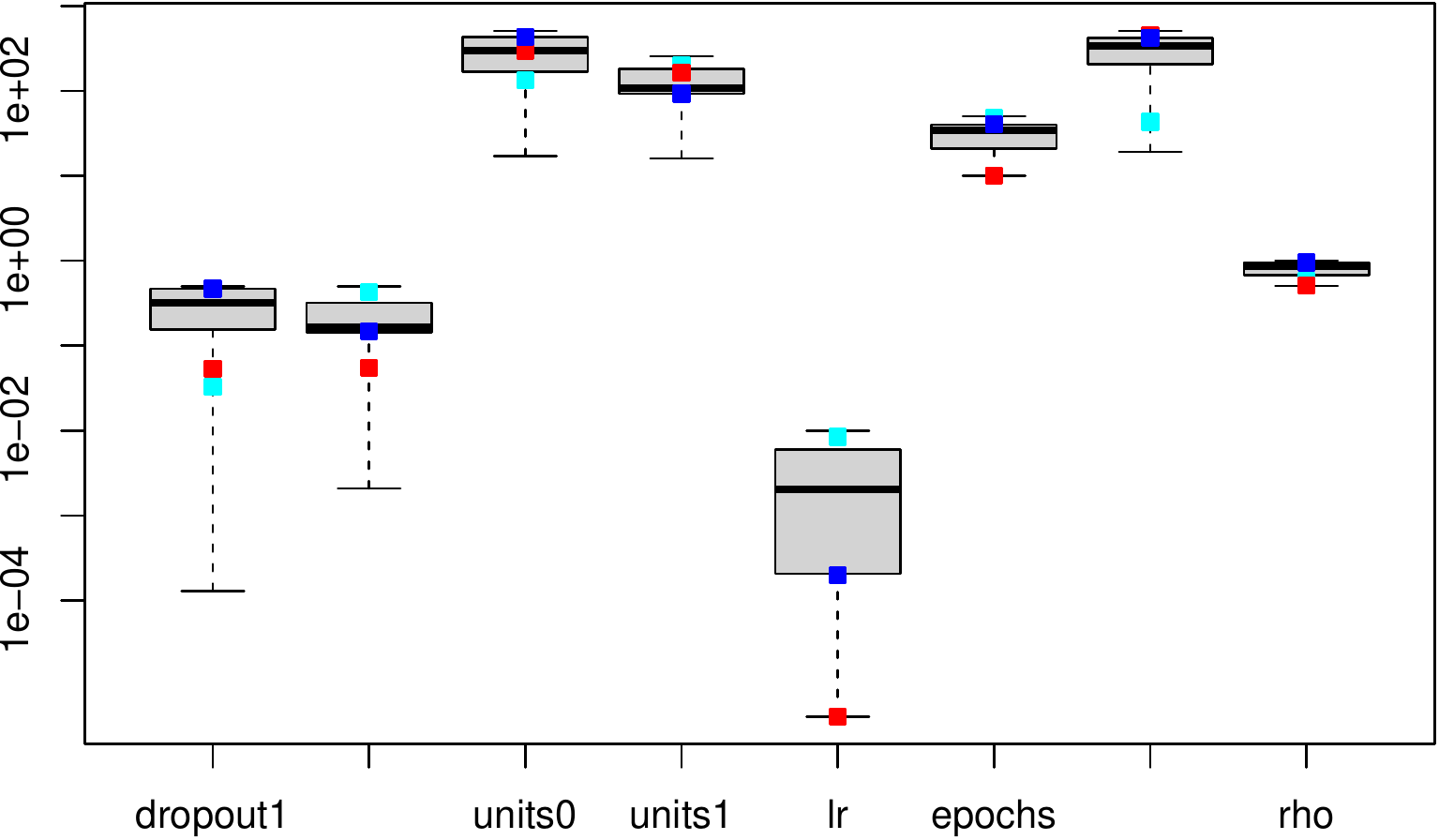}
\caption{\label{fig:boxplots}Eight box plots, i.e., each plot represents
the values of one parameter (plotted on a log scale). The \texttt{red}
square represent the worst value, the \texttt{blue} one show the
settings of the best value, and the \texttt{cyan} one show the worst
value from the tuning phase.}
\end{figure}

\emph{Box plots.} Secondly, looking at the relationship between
interesting hyperparameter configurations from this experiment might be
insightful: therefore, Fig. \ref{fig:boxplots} visualizes (using box
plots) the ranges of the eight hyperparameters from the complete
\gls{HPT} experiment. Figure \ref{fig:boxplots} shows information about
the best hyperparameter configuration (colored in blue), the worst
configuration (red), and the worst configuration from the tuning phase
(cyan).

\begin{figure}
\centering
\includegraphics{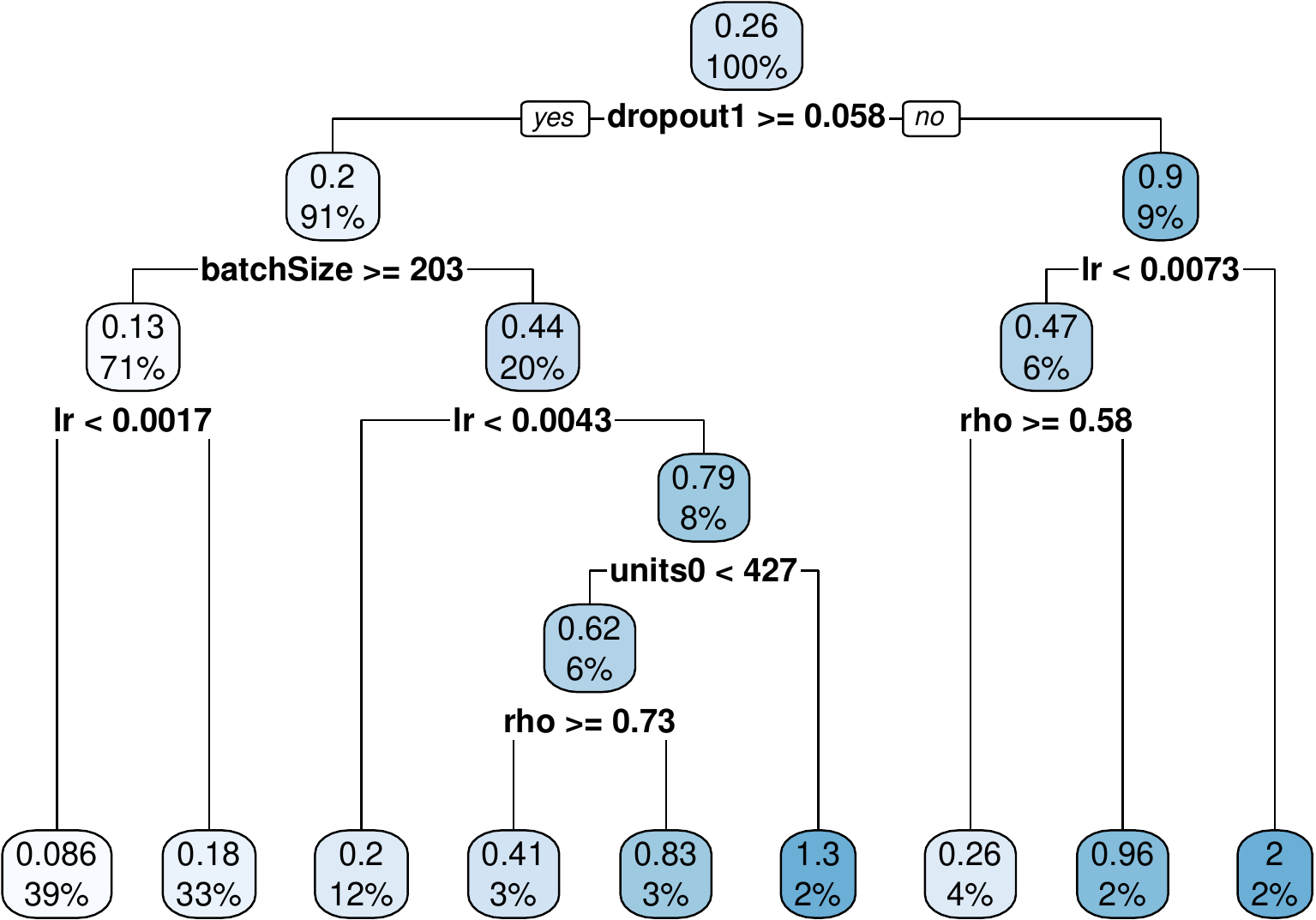}
\caption{\label{fig:fitTree}Regression tree based on the first run with
600 evaluations. Apparently, hyperparameter values from \(x_1\), \(x_5\)
and \(x_7\) are important. This result supports the previous analysis.}
\end{figure}

\begin{figure}
\centering
\includegraphics{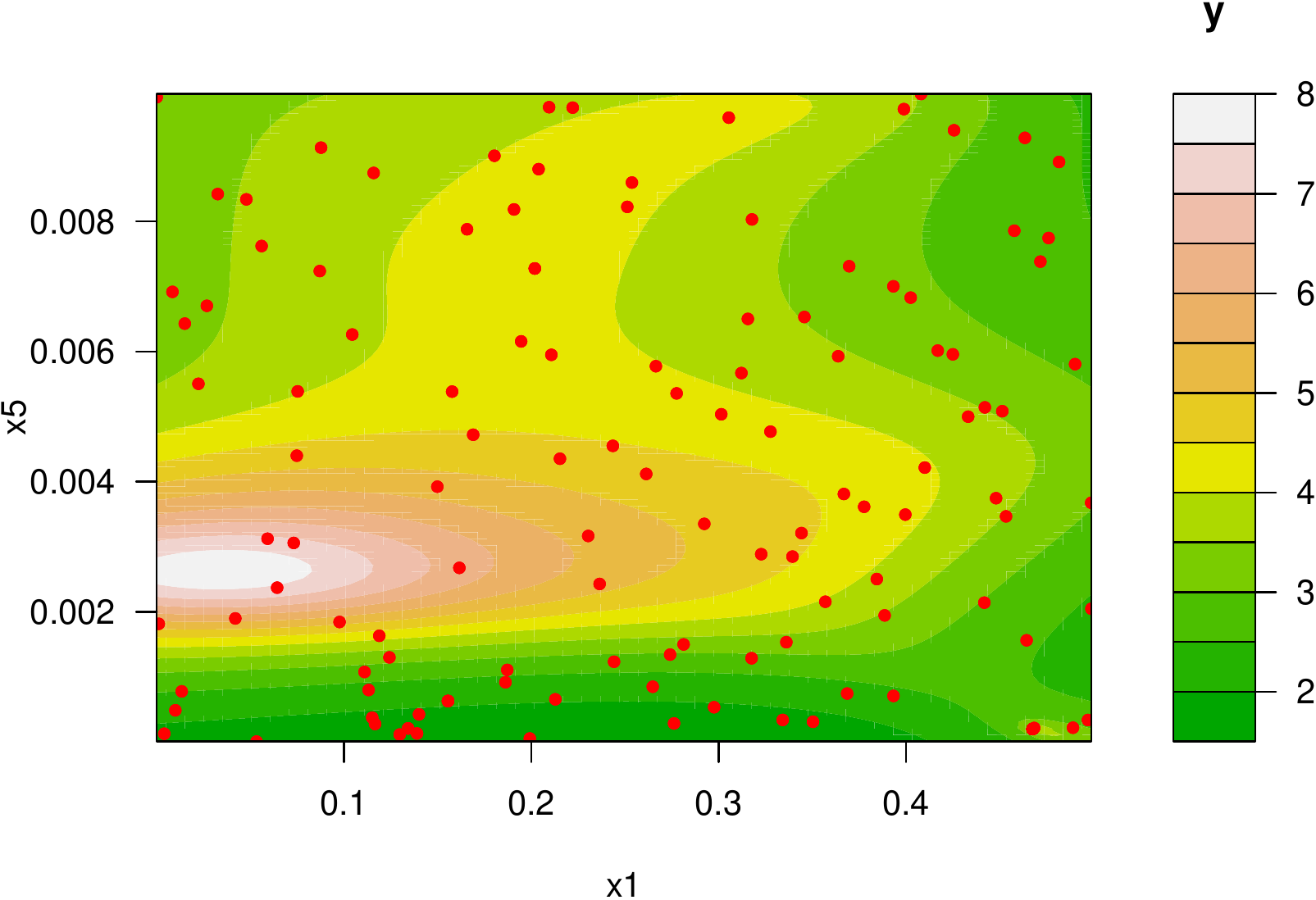}
\caption{\label{fig:oneVsFive}Surface plot: learning rate \(x_5\)
plotted against dropout1 \(x_1\).}
\end{figure}

\emph{Regression trees.} Thirdly, to analyze effects and interactions
between hyperparameters, a simple regression tree can as shown in Fig.
\ref{fig:fitTree} can be used. The regression tree supports the
observations, that hyperparameter values for \(x_1\), i.e., the dropout
rate (first layer), \(x_5\), i.e., the learning rate, and \(x_7\), i.e.,
the batch size are relevant. To conclude this first analysis,
interactions will be visualized. \gls{SPOT} provides several tools for
the analysis of interactions. Highly recommended is the use of contour
plots as shown in Fig. \ref{fig:oneVsFive}.

Figure \ref{fig:oneVsFive} supports the observations, that
hyperparameters \(x_1\) and \(x_5\) have significant effects on the loss
function.

\begin{Shaded}
\begin{Highlighting}[]
\FunctionTok{summary}\NormalTok{(result}\SpecialCharTok{$}\NormalTok{y)}
\end{Highlighting}
\end{Shaded}

\begin{verbatim}
##        V1         
##  Min.   :0.06850  
##  1st Qu.:0.08563  
##  Median :0.12405  
##  Mean   :0.25914  
##  3rd Qu.:0.21495  
##  Max.   :3.26130
\end{verbatim}

\emph{Linear models.} Finally, a simple linear regression model can be
fitted to the data. Based on the data from \gls{SPOT}'s \texttt{res}
list, this can be done as follows:

\begin{Shaded}
\begin{Highlighting}[]
\NormalTok{lm.res }\OtherTok{\textless{}{-}} \FunctionTok{lm}\NormalTok{(res}\SpecialCharTok{$}\NormalTok{y }\SpecialCharTok{\textasciitilde{}}\NormalTok{ res}\SpecialCharTok{$}\NormalTok{x)}
\FunctionTok{summary}\NormalTok{(lm.res)}
\end{Highlighting}
\end{Shaded}

\begin{verbatim}
## 
## Call:
## lm(formula = res$y ~ res$x)
## 
## Residuals:
##      Min       1Q   Median       3Q      Max 
## -0.56771 -0.14752 -0.01971  0.11414  2.13325 
## 
## Coefficients:
##               Estimate Std. Error t value Pr(>|t|)    
## (Intercept)  6.687e-01  1.005e-01   6.653 7.97e-11 ***
## res$x1      -4.783e-01  9.330e-02  -5.126 4.32e-07 ***
## res$x2       2.856e-01  1.007e-01   2.837  0.00475 ** 
## res$x3       4.506e-04  1.034e-04   4.357 1.62e-05 ***
## res$x4      -1.798e-04  2.181e-04  -0.824  0.41010    
## res$x5       4.716e+01  4.353e+00  10.832  < 2e-16 ***
## res$x6       7.589e-03  1.224e-03   6.201 1.23e-09 ***
## res$x7      -8.828e-04  9.885e-05  -8.931  < 2e-16 ***
## res$x8      -6.831e-01  9.154e-02  -7.462 4.14e-13 ***
## ---
## Signif. codes:  0 '***' 0.001 '**' 0.01 '*' 0.05 '.' 0.1 ' ' 1
## 
## Residual standard error: 0.2763 on 471 degrees of freedom
## Multiple R-squared:  0.5149, Adjusted R-squared:  0.5066 
## F-statistic: 62.49 on 8 and 471 DF,  p-value: < 2.2e-16
\end{verbatim}

Although this linear model requires a detailed investigation (a
mispecification analysis is necessary) it also is in accordance with
previous observations that hyperparameters \(x_1\), \(x_5\) and \(x_7\)
(and in addition to the previous observations, also \(x_8\)) have
significant effects on the loss function.

\section{Discussion and Conclusions}\label{sec:discussion}

This study briefly explains, how \gls{HPT} can be used as a datascope
for the optimization of \gls{DNN} hyperparameters. The results and
observations presented in Sec.\ref{sec:results} can be stated as
hypotheses, e.g.,

\begin{description}
\item[(H-1):]
hyperparameter  $x_1$, i.e.,  the dropout rate,
has a significant effect on the loss function. Its values should larger than zero.
\end{description}

This hypothesis requires further investigations. The results scratch on
the surface of the \gls{HPT} set of tools, e.g., the role and and impact
of noise was not considered. \gls{SPOT} provides very powerful tools
such as \gls{OCBA} to handle noisy function evaluations efficiently
\citep{Chen97a} \citep{Bart11b}.

Furthermore, there seems to be an upper limit for the values of the loss
function: no loss function values are larger than 3.

Considering the research goals stated in Sec. \ref{sec:introduction},
the \gls{HPT} approach presented in this study provides many tools and
solutions. Wheres in \gls{ML} and optimization, standard workflows are
available, e.g., \gls{CRISP-DM} and \gls{DOE}, the situation in \gls{DL}
is different. It might take some time until a \gls{CRISP-DL} will be
established, because several, fundamental questions are not fully
answered today.

In addition to the research goals (R-1) to (R-8) from Sec.
\ref{sec:introduction}, important goals that are specific for \gls{HPT}
in \gls{DNN} should be mentioned. We will discuss problem and algorithm
designs separately:

\begin{description}
\item[Problem Design.]
The \emph{problem design} comprehends the set of parameters that related to the 
problem. In \gls{HPT} and \gls{HPO},  regression or classification tasks  
are often considered. In our study, the \gls{MNIST} data set was chosen.
\begin{itemize}
\item Selection of an adequate performance measure: \citet{Kedz20a} claimed that \enquote{research strands into \gls{ML} performance evaluation remain arguably disorganised, $[\ldots]$. Typical \gls{ML} benchmarks focus on minimising both loss functions and processing times, which do not necessarily encapsulate the entirety of human requirement.}
\item A sound test problem specification is necessary, i.e., 
train, validation, and test sets should be clearly specified.
\item Initialization (this is similar to the specification of starting points in optimization) procedures should be made transparent.
\item Usage of surrogate benchmarks should be considered (this is similar to the use of CFD simulations in optimization)
\item Repeats (power of the test, severity), i.e., how many runs are feasible or necessary?
\item What are meaningful differences (w.r.t. specification of the loss function or accuracy)?
\item Remember: scientific relevance is not identical to statistical significance.
\item Floor and ceiling effects should be avoided.
\item Comparison to baseline (random search, random sampling, mean value $\ldots$) is a must.
\end{itemize}
\item[Algorithm Design.]
The \emph{algorithm design} in \gls{HPT} and \gls{HPO} refers to the 
model, i.e., \glspl{DNN}.
In our study, the neural network from Sec. \ref{sec:nn} was chosen.
\begin{itemize}
\item A sound algorithm (neural network) specification us required.
\item Initialization, pre-training (starting points in optimization). Pre-tuning should be explained.
\item Hyperparameter (ranges, types) should be clearly specified.
\item Are there any additional (untunable) parameters?
\item How is  noise (randomness, stochasticity) treated?
\item How is reproducibility ensured (and by whom)?
\item Last but not least: open source code and open data should be provided. 
\end{itemize}
\end{description}

To conclude: differences between \gls{HPT} and \gls{HPO} were discussed.
A \gls{HPT} approach based on \gls{SMBO} was introduced and exemplified.
It combines two packages from the statistical programming environment
\gls{R}: \gls{tfruns} and \gls{SPOT}, hence providing a \gls{HPT}
environment that is fully accessible from \gls{R}.

Although \gls{HPT} can be performed with \gls{R} functions, the
underlying \gls{TF} (Keras) and Python environment has be be installed.
This installation is explained in the Appendix.

\section{Appendix}\label{sec:appendix}
\subsection{Software Installations}
\subsubsection{Installing Python}
\paragraph{Create and Activate a Python Environment in the Project}

It is recommended that one Python virtual environment is used per
experiment. Navigate into the project directory by using the following
command:

\begin{Shaded}
\begin{Highlighting}[]
\NormalTok{cd }\SpecialCharTok{\textless{}}\NormalTok{project}\SpecialCharTok{{-}}\NormalTok{dir}\SpecialCharTok{\textgreater{}}
\end{Highlighting}
\end{Shaded}

Create a new virtual environment in a folder called \texttt{python}
within the project directory using the following command:

\begin{Shaded}
\begin{Highlighting}[]
\NormalTok{virtualenv python}
\end{Highlighting}
\end{Shaded}

The virtualenv can be activated using the following command in a
terminal:

\begin{Shaded}
\begin{Highlighting}[]
\NormalTok{source python}\SpecialCharTok{/}\NormalTok{bin}\SpecialCharTok{/}\NormalTok{activate}
\end{Highlighting}
\end{Shaded}

To verify that the correct version of Python was activated the following
command can be executed in a terminal:

\begin{Shaded}
\begin{Highlighting}[]
\NormalTok{which python}
\end{Highlighting}
\end{Shaded}

\paragraph{Install Python packages in the Environment}

Python packages such as numpy, pandas, matplotlib, and other packages
can be installed in the Python virtualenv by using
\texttt{pip\ install}:

\begin{Shaded}
\begin{Highlighting}[]
\NormalTok{pip install numpy pandas matplotlib tensorflow}
\end{Highlighting}
\end{Shaded}

\paragraph{Install and Configure reticulate to use the Correct Python Version}

Install the \texttt{reticulate} package using the following command in
the R console (e.g., from within RStudio):

\begin{Shaded}
\begin{Highlighting}[]
\FunctionTok{install.packages}\NormalTok{(}\StringTok{"reticulate"}\NormalTok{)}
\end{Highlighting}
\end{Shaded}

To configure \texttt{reticulate} to point to the Python executable in
the virtualenv \texttt{python} from above, create a file in the project
directory called \texttt{.Rprofile} with the following contents:

\begin{Shaded}
\begin{Highlighting}[]
\FunctionTok{Sys.setenv}\NormalTok{(}\AttributeTok{RETICULATE\_PYTHON =} \StringTok{"python/bin/python"}\NormalTok{)}
\end{Highlighting}
\end{Shaded}

\gls{R} (or RStudio) must be restarted for the setting to take effect.
To check that \texttt{reticulate} is configured for the correct version
of Python the following command can be used in the R (or RStudio)
console:

\begin{Shaded}
\begin{Highlighting}[]
\NormalTok{reticulate}\SpecialCharTok{::}\FunctionTok{py\_config}\NormalTok{()}
\end{Highlighting}
\end{Shaded}

\subsubsection{Installing Keras}

To get started with Keras, the Keras \gls{R} package, the core Keras
library, and a backend tensor engine (such as \gls{TF}) must be
installed. This can be done as follows from within R (or RStudio):

\begin{Shaded}
\begin{Highlighting}[]
\FunctionTok{install.packages}\NormalTok{(}\StringTok{"tensorflow"}\NormalTok{)}
\FunctionTok{install.packages}\NormalTok{(}\StringTok{"keras"}\NormalTok{)}
\FunctionTok{library}\NormalTok{(keras)}
\FunctionTok{install\_keras}\NormalTok{()}
\end{Highlighting}
\end{Shaded}

This will provide you with a default CPU-based installation of Keras and
\gls{TF}. To install a GPU-based version of the \gls{TF} backend engine,
the corresponding command reads as follows:

\begin{Shaded}
\begin{Highlighting}[]
\FunctionTok{install\_keras}\NormalTok{(}\AttributeTok{tensorflow =} \StringTok{"gpu"}\NormalTok{)}
\end{Highlighting}
\end{Shaded}

\subsubsection{Installing SPOT}

The following commands can be used to install the most recent version of
\gls{SPOT} and the additional package \gls{SPOTMisc} from \gls{CRAN}:

\begin{Shaded}
\begin{Highlighting}[]
\FunctionTok{install.packages}\NormalTok{(}\StringTok{"SPOT"}\NormalTok{)}
\FunctionTok{install.packages}\NormalTok{(}\StringTok{"SPOTMisc"}\NormalTok{)}
\end{Highlighting}
\end{Shaded}

Further information about the most recent \gls{SPOT} versions will be
published on \url{https://www.spotseven.de/spot/}.

\subsection{The Hyperparameter Optimization Problem}

For convenience, we include the definitions of
\enquote{learning algorithms} and \enquote{hyperparameter optimization}
from \cite{Berg12a}. The objective of a learning algorithm \(A\) is to
find a function \(f\) that minimizes some expected loss \(L(x; f)\) over
i.i.d. samples \(x\) from a natural (ground truth) distribution \(G_x\).

\begin{defn}[Learning algorithm; \citep{Berg12a}]
A \emph{learning algorithm}\/ $A$ is a functional 
that maps a data set $X^{(\text{train})}$ 
(a finite set of samples from $G_x$) 
to a function $f$. 
\end{defn}

A learning algorithm can estimate \(f\) through the optimization of a
training criterion with respect to a set of parameters \(\theta\). The
learning algorithm itself often has hyperparameters \(\lambda\), and the
actual learning algorithm is the one obtained after choosing
\(\lambda\), which can be denoted \(A_{\lambda}\), and
\(f = A_{\lambda}(X^{(\text{train})})\) for a training set
\(X(^\text{train})\).

Practitioners are interested in a way to choose \(\lambda\) so as to
minimize generalization error, which can be defined as follows.

\begin{defn}[Generalization error; \citep{Berg12a}]
\begin{equation}
E_{x \sim G_x} \left[ L(x;A_{\lambda}(X^{(\text{train})})) \right].
\end{equation}
\end{defn}

The computation performed by \(A\) itself often involves an inner
optimization problem (optimizing the weights of a \gls{NN}), which is
usually iterative and approximate.

The problem of identifying a good value for hyperparameters \(\lambda\)
is called the problem of hyperparameter optimization. The outer-loop
optimization problem, which is of great practical importance in
empirical machine learning work, can be stated as follows:
\begin{equation}
\lambda^{(*)} = \argmin_{\lambda \in \Lambda} E_{x \sim G_x} \left[ L\left( x; A_{\lambda} (X^{(\text{train})} ) \right) \right].
\end{equation}

\begin{defn}[Hyperparameter optimization problem; \citep{Berg12a}]
The \emph{hyperparameter optimization problem}  can be stated 
in terms of a hyperparameter response function, $\Psi$ as follows:
\begin{align}
\lambda^{(*)} \approx & \argmin_{\lambda \in \Lambda} \frac{1}{| \Xval|} \sum_{x \in  \Xval} L \left( x; A_{\lambda} (\Xtrain ) \right)\\
 & = \argmin_{\lambda \in \Lambda} \Psi(\lambda) \\\label{eq:hatlambda}
 & \approx \argmin_{\{ \lambda^{(i)} \}_{i = 1,2, \ldots, S}} \Psi(\lambda) = \hatlambda.
\end{align}
\end{defn}

We can define \gls{HPO} as a minimization problem:

\begin{defn}[Hyperparameter optimization; \citep{Berg12a}]
Hyperparameter optimization is the minimization of 
\begin{equation}
\Psi ( \lambda ) \text{ over }  \lambda \in \Lambda.
\end{equation}
\end{defn}

\begin{defn}[Hyperparameter surface]\label{def:surface}
Similar to the definition in \gls{DOE}, the function $\psi \in \Psi$ is referred to as the hyperparameter \emph{response surface}.
\end{defn}

Different data sets, tasks, and learning algorithm families give rise to
different sets \(\Lambda\) and functions \(\Psi\). Knowing in general
very little about \(\Psi\) or the search space \(\Lambda\), the dominant
strategy for finding a good \(\lambda\) is to choose some number \((S)\)
of trial points \(\{ \lambda^{(i)} \}_{i = 1,2, \ldots, S}\), to
evaluate \(\Psi(\lambda)\) for each one, and return the
\(\lambda^{(i)}\) that worked the best as \(\tilde{\lambda}\). This
strategy is made explicit by Equation \ref{eq:hatlambda}.

Whereas \(\lambda\) denotes an arbitrarily chosen hyperparameter
configuration, important hyperparameter configurations will be labeled
as follows: \(\lambda_i\) is the \(i\)-th hyperparameter configuration,
\(\lambda^{(\star)}(t)\) is the best hyperparameter configuration at
iteration \(t\), and \(\lambda^{(\star)}\) is the final best
hyperparameter configuration.

\begin{defn}[Low effective dimension; \citep{Berg12a}]
\label{def:led}
If a function $f$ of two variables could be approximated by 
another function of one variable $(f(x_1,x_2) \approx g(x_1))$,
we could say that $f$ has a \emph{low effective dimension}. 
\end{defn}

\printnoidxglossaries

\bibliographystyle{abbrvnat}
\bibliography{./bart21g.bib}

\end{document}